\documentclass[pdflatex,sn-mathphys-num]{sn-jnl}% Math and Physical Sciences Numbered Reference Style
\usepackage{graphicx}%
\usepackage{multirow}%
\usepackage{amsmath,amssymb,amsfonts}%
\usepackage{amsthm}%
\usepackage{mathrsfs}%
\usepackage[title]{appendix}%
\usepackage{xcolor}%
\usepackage{textcomp}%
\usepackage{manyfoot}%
\usepackage{booktabs}%
\usepackage{algorithm}%
\usepackage{algorithmicx}%
\usepackage{algpseudocode}%
\usepackage{listings}%
\usepackage{bm}
\theoremstyle{thmstyleone}%
\newtheorem{theorem}{Theorem}%  meant for continuous numbers
\theoremstyle{thmstyletwo}%
\newtheorem{remark}{Remark}%

\theoremstyle{thmstylethree}%

\begin{document}

\title[Clustered Calibration]{Clustered Calibration: Representation-Aware Probability Calibration via Learned Subpopulations}

%%=============================================================%%
%% GivenName	-> \fnm{Joergen W.}
%% Particle	-> \spfx{van der} -> surname prefix
%% FamilyName	-> \sur{Ploeg}
%% Suffix	-> \sfx{IV}
%% \author*[1,2]{\fnm{Joergen W.} \spfx{van der} \sur{Ploeg} 
%%  \sfx{IV}}\email{iauthor@gmail.com}
%%=============================================================%%

\author*[1]{\fnm{Tomer} \sur{Lavi}}\email{tomerlav@post.bgu.ac.il}
\author*[1]{\fnm{Bracha} \sur{Shapira}}\email{bshapira@bgu.ac.il}
\author*[1]{\fnm{Nadav} \sur{Rappoport}}\email{nadavrap@bgu.ac.il}

\affil*[1]{\orgdiv{Faculty of Computer and Information Science}, \orgname{Ben-Gurion University of the Negev}, \orgaddress{\country{Israel}}}

%%==================================%%
%% Sample for unstructured abstract %%
%%==================================%%

\abstract{Ensuring that predicted probabilities align with observed frequencies is critical in high-stakes domains such as clinical decision support, autonomous driving and financial risk assessment. Existing calibration methods typically apply a single global transformation or rely on post-hoc binning over predicted confidences, limiting their ability to exploit heterogeneous reliability across sub-populations. We propose Clustered Calibration, a representation-aware framework that identifies sub-populations via clustering in learned feature spaces (e.g., coverage vectors, SHAP values, CNN activations, Transformer embeddings) and fits a soft mixture of cluster-specific parametric calibrators under hierarchical shrinkage toward a global mapping. This design yields context-specific calibration while maintaining global stability. Across six tabular datasets and additional image and text benchmarks, clustered calibration consistently improves or matches strong global calibrators in terms of negative log-likelihood and Brier score, while preserving AUC and accuracy. We further show, both analytically and empirically, that fixed-bin Expected Calibration Error (ECE) can mis-rank soft, region-aware calibrators even when proper scoring rules improve, and we advocate for log-loss and Brier as more reliable bases for model selection in such settings.}


%%\pacs[JEL Classification]{D8, H51}

%%\pacs[MSC Classification]{35A01, 65L10, 65L12, 65L20, 65L70}

\maketitle

\section{Introduction}
When a machine learning model is involved in high-stakes domains e.g. clinical decision support systems (CDSS) in healthcare, autonomous driving and financial risk assessment, it is critical that its predictions are calibrated as well as accurate \cite{minderer2021revisiting}. Inaccurate outcome probability estimates might lead to hazardous results \cite{staartjes2020importance}.
Global calibration methods apply a uniform transformation across all predictions, disregarding systematic variations in reliability across different subsets of data. Local calibration methods do address the need for subset-specific calibration, but they often require a-priori identification of subsets or still rely on a unified global calibration function, which may not fully capture the unique characteristics of each subgroup. An alternative approach is binning, where predictions are partitioned into discrete intervals by their predicted probabilities, and calibration parameters are learned within each bin. Recent extensions combine parametric calibration with binning or local neighborhoods, yet they fail to fully harness the latent geometry of the data and the model’s internal representations in order to enable more fine‑grained and stable calibration without requiring predefined subgroups.
Binning is also used as a basis for the Expected Calibration Error metric (ECE). Bins, defined post-hoc based on model outputs rather than the underlying data structure, can miss meaningful pattern discontinuities. While binning can partially account for heterogeneity in reliability, it potentially ignores structure in the input space that may include heterogeneous subpopulations. In Section \ref{sec:ECE-mis-rank}, we show that fixed-bin ECE can prefer a worse probabilistic model when calibration is local/soft, while proper scores, e.g. log-loss and Brier, improve.

In this work, we propose Clustered Calibration, a cluster-based framework that leverages data-driven subpopulation discovery to perform heterogeneous calibration while maintaining global calibration.
Modern machine learning models inherently learn latent representations that partition the input space. These latent spaces capture regions of similar predictive patterns, often encoding relationships that are not directly observable from raw predicted probabilities. Leveraging them for binning enables the creation of clusters that group together samples with similar model behavior, regardless of their raw output. This cluster‑aware approach can expose systematic differences in calibration across regions of the feature space, allowing for finer‑grained and more accurate adjustments than output‑based binning alone. 
Beyond improved calibration, clustering offers additional advantages: it provides an efficient data representation, since only cluster centers are stored and used, reducing computational cost compared to instance-based methods such as K‑Nearest Neighbors. It also enhances explainability by identifying coherent sub‑populations as clusters, aiding in the interpretation and communication of model behavior.

Our framework is model and domain agnostic. We use tabular datasets as our primary case study to demonstrate these benefits, in addition to the image and text domains. Tabular data not only allow for clear interpretation of cluster structure through typical feature values within each cluster, linking subpopulations to meaningful domain concepts, but also cover a wide range of real-world applications such as healthcare, finance, and risk assessment where calibration quality has immediate practical consequences. We also evaluate our method on image and text classifiers to assess its behavior in deep representation spaces and to demonstrate that the same clustered calibration framework extends beyond tabular models. As Section~\ref{experimental-results} demonstrates, tabular data yields the most consistent and substantial improvements over strong global calibrators. This tabular setting therefore serves as the central empirical case study of our work, with image and text experiments used to probe robustness in deep representation spaces.

The main contributions of this work are as follows: (1) we propose a cluster‑based soft binning framework that leverages latent representations to form sub‑populations for calibration, (2) we incorporate soft cluster membership weights and L2‑shrinkage toward global parameters to improve both global and local calibration and (3) we empirically demonstrate strong performance across diverse datasets compared with strong calibration baselines.

\textbf{Main claim and evidence.} Our main claim is that a soft, representation-aware clustered calibration scheme can improve global and local calibration measured by proper scoring rules such as negative log-likelihood and Brier score, without sacrificing discrimination. Our method exploits sub-population structure in learned representations (coverage vectors, SHAP values, CNN activations and Transformer embeddings). We substantiate this claim through an extensive empirical study on six tabular datasets and a variety of image and text benchmarks, where clustered calibration consistently improves or matches strong global parametric calibrators in log-loss and Brier score, achieves competitive AUC and accuracy, and yields interpretable clusters that align with meaningful sub-populations. Additionally, we provide a theoretical and empirical analysis showing that fixed-bin Expected Calibration Error (ECE) can mis-rank calibrated predictors in soft, region-aware settings, whereas proper scoring rules remain aligned with improved probabilistic predictions.

\section{Background and Related Work}
\subsection{Calibration Function}
\begin{figure}
    \centering
    \includegraphics[width=1\linewidth]{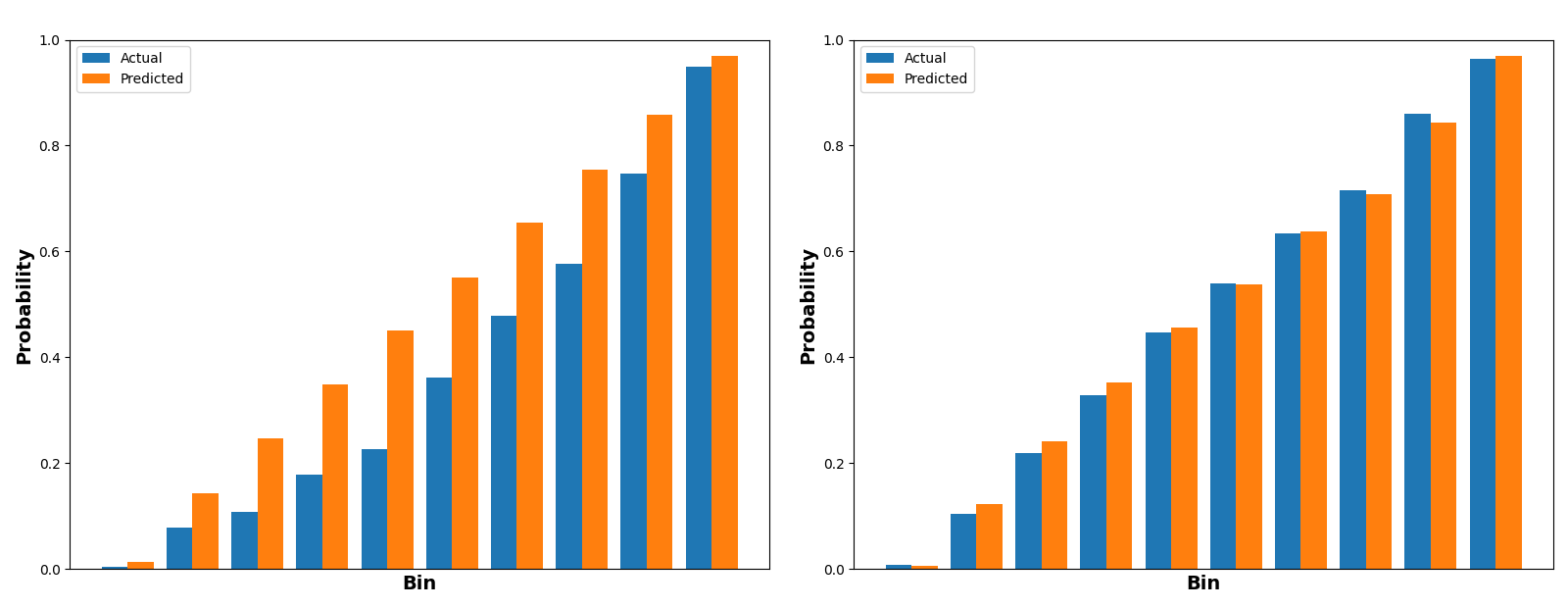}
    \caption{A classification model calibration comparison. The X-axis shows predictions binned by model confidence (10 equal-width bins, and the Y-axis is the mean probability of the predicted and observed values. The right diagram displays the calibrated outputs of the left diagram. The orange bars are the predicted probabilities, the blue bars are the actual frequencies.} 
    \label{fig:Calibration}
\end{figure}

Since the underlying assumption of a machine learning-based classification is that the model’s output represents the frequency of the predicted event, we can define the model's predicted probability to be \(\hat{p}(y|x)\). Many classification models are over-confident in their output and the predicted probability tends to be significantly higher than the actual frequency of the labels. A calibration method, $c$, transforms the predicted score of the model \(\hat{p}\) into a calibrated probability \(p\), which is expected to be better aligned with the observed frequency, as follows: \(c(\hat{p},x) = p(y|\hat{p},x)\).
Figure \ref{fig:Calibration} displays the output scores of a classification model before and after calibration. Left: Uncalibrated classifier reliability diagram. Right: Calibrated output, which is significantly closer to the actual frequency of the target labels along the bins.

\subsection{Calibration Error Metrics}
\label{Sec:CalibrationErrorMetrics}
The quality of a model's calibration is measured by the difference between its predicted probability of test samples and their observed frequency. 
Expected Calibration Error (ECE) \cite{naeini2015obtaining} is defined for classification as follows: 
\begin{equation}
\label{eq:ECE}
ECE = \frac{1}{N}\overset{M}{\underset{i=1}{\sum}}|B_i|\cdot \lvert A(B_i)-\hat{P}(B_i) \rvert
\end{equation}
Where \(N\) is the total number of test samples, \(B = \{B_1,...,B_M\}\) a division of the samples sorted by their predicted probabilities to M bins. \(B_i\) contains samples with output range \([\frac{i-1}{M}, \frac{i}{M}]\), \(A\) is the true proportion of positive samples, and \(P\) is the average output probability of the calibration method. \textit{Maximum Calibration Error (MCE)} measures the maximum error over the bins: 
\[MCE = \max_{i=1...M} \lvert A(B_i)-\hat{P}(B_i) \rvert\]
\textit{Adaptive ECE (AdaECE)} divides the samples into equal-sized bins \cite{nguyen2015posterior}. AdaECE uses the Brier score as the error term  \cite{brier1950verification} , and bins are split using the adaptive binning method s.t. the samples are distributed equally among the bins. AdaECE is calculated as the square root of the sum over the bins of the difference between the predicted and observed probabilities:
\[AdaECE = \sqrt{\frac{1}{N}\overset{M}{\underset{i=1}{\sum}}|B_i|\cdot  (P(B_i)-\hat{P}(B_i))^2
}\]
AdaECE is claimed to be more reliable than density-based ECE in which only a small volume of samples falls into the mid-range bins. In this paper, we will be using AdaECE as our default metric (referred to as ECE from here onward).

\subsection{Calibration Methods}
\subsubsection{Global Calibration}
Platt Scaling is considered a baseline machine learning calibration method \cite{Platt2000}. It is based on learning a linear transformation of the model's logit values. The learned parameters are then used to transform the output of the model to the calibrated values. 
\[p(y=1|f, x) = \frac{1}{1 + exp(A \cdot f(x) + B)}\]
Where $f(x)$ is the output logits of the prediction model and A, B are learned scalars. Vector Scaling is the multi-class version of Platt scaling \cite{guo2017calibration}.

Other known parametric methods are Temperature Scaling (TS) \cite{guo2017calibration}, Beta and its multi-class variant Dirichlet calibration \cite{kull2017beyond}, \cite{kull2019beyond}.
These methods apply a unified global transformation on the predicted values, thus are not consistent in the error they impose over the range of the samples. 

\subsubsection{Score-based binning methods}
A separate category of calibration methods partitions the samples into bins and computes a calibrated output per bin. Histogram Binning \cite{zadrozny2001obtaining} divides the sample space into bins of probability scores and assigns a calibrated score per bin. Isotonic Regression extends Histogram Binning by learning bin boundaries \cite{zadrozny2002transforming}. Bayesian Binning into Quantiles (BBQ) divides the data into bins by using a Bayesian model \cite{naeini2015obtaining}. Separating the outputs into bins aims to differentiate the samples, but the methods use a na\"{\i}ve  partitioning approach, ignoring the variety of sub-distributions in the sample space and therefore yields inconsistent calibration across subpopulations.

Platt-Bin and Scaling-Binning are a combination of the Platt scaling and Histogram Binning methods, where scaling is performed prior to obtaining the Histogram Binning method \cite{singh2022platt, kumar2019verified}. 
Confidence-based Temperature scaling  (CBT) \cite{frenkel2022network} and Bin-wise Temperature Scaling (BTS) \cite{ji2019bin} further develop the composition to an ensemble in which TS is applied to each bin individually, thus improving the overall calibration. Another combined approach is Mix-n-Match which creates a generic calibration model which is composed of an ensemble of weighted calibration functions that are mutually trained on the model's output \cite{zhang2020mix}.

Although these methods assume different probability distributions across the bins, they are limited by the probability output values as their input and therefore struggle to capture of the distribution of subpopulations. 

\subsubsection{Multicalibration}
Several studies identified the need to consider the probability distribution of subpopulations of the data. Calculating a TS calibration per class in a multi-class model is proposed in \cite{frenkel2021network}. One may consider each class as an identified subpopulation, yet this method still overlooks subpopulations within classes, and is not applicable to binary classification tasks. Dirichlet local calibration is the multivariate version of the Beta calibration based on the Dirichlet probability distribution of class labels.

Multicalibration requires that predictions are accurate not only across the entire sample space, but also within any subpopulation defined by a specified computational class (e.g. decision trees, boolean features) \cite{hebert2018multicalibration}.  Since multicalibration training depends on the pre-identified subpopulations, it might overlook groups which were not chosen to be optimally calibrated. Our goal is to develop a method that complements multicalibration by identifying subpopulations that are implicitly learned by the prediction model and may have been overlooked. This requires exploration of the mapping of the sample space as it is learned by the predictor. 

\subsubsection{Local Calibration}
Local Calibration Error (LCE) and Maximum Local Calibration Error (MLCE) \cite{luo2022local} address the requirement to measure calibration on individual levels without prior knowledge on subsets of the population. It defines a soft local neighborhood for each point and calculates a confidence based calibration over that neighborhood. Thus, it captures fine-grained, context-specific mis-calibration that global ECE can miss. Local Recalibration (LoRe) is the accompanied recalibration method which is based on calculating calibrated score in the neighborhood of each point based on the kernel similarity and its assigned bin.

\paragraph{ECE Limitations on Local Calibration}
\label{subsubsection:ece_limitations}
LCE and MLCE address the limitations of ECE in measuring local calibration: ECE's bin-level aggregation obscures fine-grained, context-specific miscalibration patterns. Additionally, bin construction is inherently arbitrary: the choice of bin count, binning scheme (equal width or quantile-based) and bin boundaries critically influences the resulting error estimate. This arbitrariness creates a circularity: when binning methods train calibration parameters within their chosen bins and then evaluate calibration on those same bins, they are effectively evaluating their own optimization target.
Recent work has further identified additional concerns about ECE's statistical properties \cite{kumar2019verified, rossellini2025can, roelofs2022mitigating}. 
Alternative evaluation frameworks based on proper scoring rules (e.g., log-loss, Brier score \cite{murphy1973new}) are method-agnostic, theoretically grounded in decision theory, and do not suffer from the circularity problem \cite{gneiting2007strictly}. They provide fair comparison across diverse calibration approaches including soft neighborhood and local methods. 
Consequently, we prioritize log-loss and Brier score as primary evaluation measures, while using ECE primarily for diagnostic purposes when comparing with hard-partition baselines.

\subsubsection{Relationship to Local Recalibration and Multicalibration}
Our approach is closely related to Local Recalibration (LoRe) and the associated Local Calibration Error (LCE) framework, but differs in several important ways. LoRe constructs a per-point local neighborhood in confidence space and computes recalibrated probabilities via kernel-weighted estimates over nearby predictions, with neighborhoods defined directly on the model’s output scores. This yields highly flexible, point-wise local recalibration but comes at a substantial inference time cost, since a potentially large set of neighbors must be consulted for every new prediction. LoRe/LCE do not expose a stable set of sub-populations that can be directly inspected or reused. In contrast, our Clustered Calibration operates in a representation space, where we first extract learned embeddings (coverage vectors, SHAP values, CNN activations or Transformer hidden states) and then partition this space into a small number of centroids via clustering. Soft memberships are computed with respect to these centroids, so the cost of calibration scales only with the number of clusters rather than the size of the calibration set. Moreover, our cluster-specific calibrators are regularized via hierarchical shrinkage toward a global mapping which explicitly trades off local adaptation against global stability and yields interpretable, persistent sub-populations whose typical feature profiles can be examined (e.g., in the Stroke case study, see: \ref{example:CDSS}).

Multicalibration takes a complementary perspective by specifying a rich family of subpopulations in advance (e.g., Boolean combinations of features or decision-tree leaves) and enforcing approximate calibration on all groups in this family, yielding strong worst-case guarantees over the chosen concept class. However, this requires a practitioner to define or choose the subgroup family a-priori, and can be computationally demanding when the class is large. Our method instead discovers subpopulations implicitly from the model’s learned representations, without requiring predefined groups, and optimizes proper scoring rules (log-loss and Brier) at both global and cluster levels under shrinkage. As a result, we do not claim multicalibration guarantees with respect to a broad concept class, but rather, we target the most salient heterogeneity that the predictor has already encoded in its internal geometry and provide a lightweight, interpretable mechanism for adapting calibration to these representation-defined subgroups. In this sense, Clustered Calibration can be viewed as complementary to multicalibration: it trades the latter’s worst-case theoretical guarantees for a data-driven, representation-aware notion of subpopulation discovery that is easy to deploy and inspect in practice.

\subsubsection{Clustering of Learned Representations}
In developing a calibration method that is aimed at capturing fine‑grained structure within the input space, we draw inspiration from prior research on clustering in transformed feature spaces. These works demonstrate how learned or engineered representations can reveal latent structure in the original input or output domains, enabling the formation of meaningful sub‑populations for targeted model adjustments. 

Deep Clustering Network (DCN) uses a combination of a learned dimension-reduced data representation to improve clustering \cite{yang2017towards}. Deep Embedded Clustering (DEC) clusters a learned representation \cite{xie2016unsupervised}. DeepSafe clusters output activations of deep neural networks layers to identify uncertain regions \cite{gopinath2017deepsafe}. 

Building on this idea, our approach uses clustering to identify and calibrate regions of the feature space that share similar predictive behavior. This design combines the benefits of localized calibration, computational tractability, and explanatory insight.

\subsubsection{Positioning}
In summary, our approach lies between global parametric calibration and multicalibration. We do not attempt to enforce calibration on an exponentially large family of subgroups. Instead, we exploit the model’s own learned representations to discover a small number of salient sub-populations via clustering, then fit shrinkage-regularized cluster-specific calibrators. Compared to binwise methods that operate purely on predicted probabilities, this representation-aware design is capable of capturing heterogeneity in reliability that is not visible at the level of scalar confidence alone, while remaining computationally lightweight and easy to inspect in practice.

\section{Method}
\label{sec:method}
\begin{figure}
    \centering
    \includegraphics[width=1\linewidth]{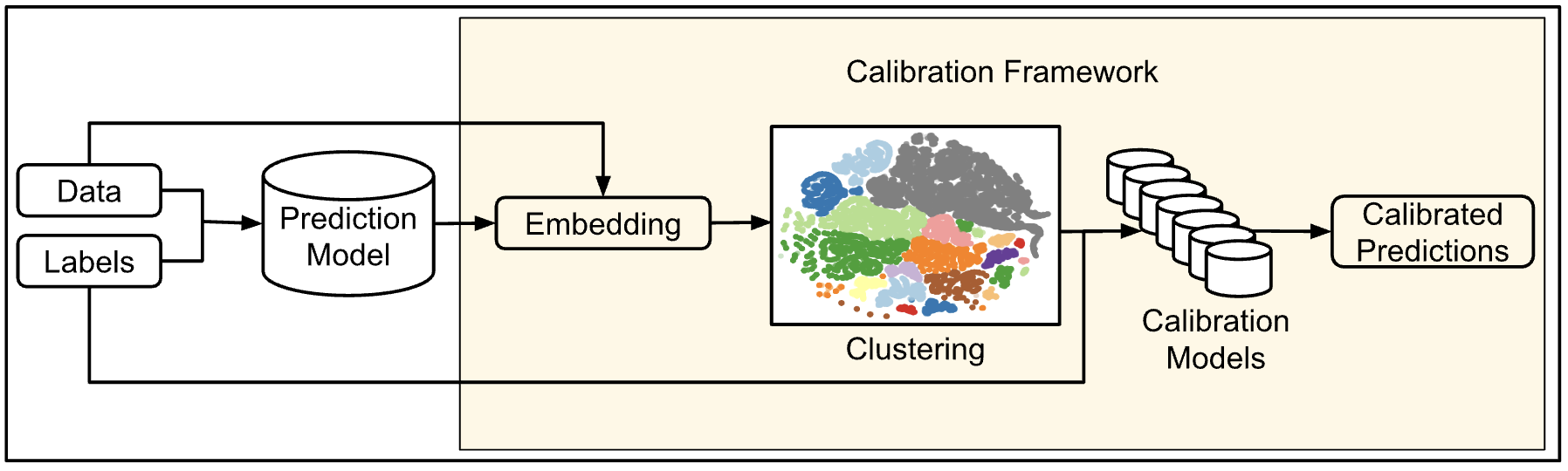}
    \caption{The architecture of the proposed clustered-calibration method.}
    \label{fig:clustered calibration architecture}
\end{figure}

\textbf{Overview}. Figure \ref{fig:clustered calibration architecture} summarizes the clustered calibration pipeline. Given a trained classifier, we first extract a learned representation $\phi(x)$ (coverage vectors or SHAP values for tree ensembles, CNN activations for images, Transformer hidden states for text). We then apply K-Means clustering in this representation space to obtain centroids and soft cluster memberships for each sample. On a held-out calibration set, we fit a global parametric calibrator and a collection of cluster-specific calibrators, all trained to minimize a proper scoring rule (log-loss or Brier) with an explicit hierarchical shrinkage penalty that discourages cluster models from deviating unnecessarily from the global mapping. At test time, we assign new input points to clusters via their soft memberships and produce calibrated probabilities as a weighted mixture of the cluster-specific calibrators.

\subsection{Problem setting and notation}

Let \((x_i,y_i)_{i=1}^n\) be i.i.d. samples with features \(x_i\in\mathbb{R}^d\) and labels \(y_i\in\{1..L\}\). A trained classifier produces uncalibrated probabilities \(\hat p_i=f(x_i)\in(0,1)^L\) on a held-out calibration split \(\mathcal D_{\text{cal}}\). Our goal is to learn a calibrated predictor \(g:(0,1)^L\times\mathbb{R}^m\to (0,1)^L\times\mathbb{R}\) that may depend on an auxiliary representation \(\phi(x)\in\mathbb{R}^m\) and yields calibrated probabilities \(p_i^{\star}=g(\hat p_i,\phi(x_i))\).

We employ an input-space representation \(\phi\) and unsupervised clustering over \(\phi(x)\) to discover subpopulations with similar learned features, then fit cluster-specific parametric calibrators under shrinkage toward a global calibrator and combine them with soft cluster membership weights at inference. 

\subsection{Representation of samples}
\subsubsection{Tree Ensembles}
\paragraph{Tree-coverage vectors (coverage)}
For each tree \(t\) and its leaves \(\mathcal L_t\), define a sparse binary indicator over the union of all leaves \(\mathcal L=\bigcup_t\mathcal L_t\): \(\phi_{\text{cov}}(x)[\ell]=\mathbb{1}\{x \text{ falls in leaf } \ell\}\). This results in a \(n\times |\mathcal L|\) matrix \(C\) with exactly \(T\) ones per row (one per tree). To sharpen cluster structure we use TF-IDF weighting and row \(\ell_2\) normalization:
\[
\tilde C = \mathrm{norm}_{\ell_2} \big(C\cdot \mathrm{diag}(\text{idf})\big),\quad
\text{idf}_j=\log \frac{n+1}{\text{df}_j+1}+1.
\]
We apply Truncated SVD \cite{halko2009finding} to \(\tilde C\) to obtain a dense embedding \(Z\in\mathbb{R}^{n\times r}\) (typically \(r\)= 256).

\paragraph{SHAP vectors}
SHAP values (SHapley Additive exPlanations) are used to explain the output of a machine learning model \cite{NIPS2017_7062}. Each input feature is given a score for each sample, where a positive value indicates that the feature promotes the predicted output and vice versa. An advantage of using SHAP values is that they can identify explainable subgroups within the samples. We compute per-sample SHAP values \(\psi(x)\in\mathbb{R}^d\) for the trained model. 

\subsubsection{Convolutional Neural Networks (CNNs)}
For CNN-based image classifiers, we extract for each sample the activation vector from the last convolutional block after global average pooling (the penultimate layer before the classification head) and use this representation as the input to the clustering step.

\subsubsection{Transformer-based Models}
For transformer-based text classifiers, we use the hidden state of the final transformer layer as a sentence-level embedding on which clustering is performed.

\subsection{Clustering and soft memberships}
We run a clustering method on the chosen embedding space. Let \(\{\mu_k\}_{k=1}^K\) be the learned centroids. For each sample \(x\) we assign soft memberships
\[
w_k(x)=\frac{\exp\big(-d^2(\phi(x),\mu_k)\big)}{\sum_{j=1}^K \exp\big(-d^2(\phi(x),\mu_j)\big)},
\]
where \(d^2\) is squared cosine distance in the embedding. Soft memberships reduce boundary discontinuities relative to hard cluster assignments. 

\subsection{Parametric calibrators}
We consider standard monotone, probability-preserving calibrators \(c_\theta:(0,1)^L\to(0,1)^L\):

\begin{itemize}
    \item Platt scaling: \(c_{\theta}(\hat p)=\sigma \big(A\cdot \mathrm{logit}(\hat p)+B\big), \theta=(A,B)\).
    \begin{itemize}
        \item On multi-label, Vector scaling applies an affine transform to each label's logit, followed by a sigmoid.
    \end{itemize}
    \item Temperature scaling: \(c_{\theta}(\hat p)=\sigma\big(\mathrm{logit}(\hat p)/T\big), \theta=(T>0)\).
    \begin{itemize}
        \item On multi-label, a shared-$T$ ties calibration across labels with a single temperature.
    \end{itemize}
    \item Beta calibration: \(c_{\theta}(\hat p)=\sigma \big(a\log \hat p + b\log(1-\hat p)+c\big), \theta=(a,b,c)\).
    \begin{itemize}
        \item On multi-label, Dirichlet calibration maps $\hat{p}$  via a log-linear transform on probabilities followed by a softmax.
    \end{itemize}
\end{itemize}

We first fit a global calibrator \(c_{\theta_0}\) on \(\mathcal D_{\text{cal}}\) by minimizing the negative log-likelihood (NLL). 

\subsection{Cluster-specific calibrators with shrinkage (hierarchical prior)}
For each cluster \(k\) we learn parameters \(\theta_k\) on \(\mathcal D_{\text{cal}}\) using softly weighted NLL with \(\ell_2\) shrinkage toward \(\theta_0\):

\begin{equation}
\label{eq:CCL-Loss}
\mathcal L_k(\theta_k) =
\sum_{(x_i,y_i)\in\mathcal D_{\text{cal}}} w_k(x_i)\underbrace{\ell\big(c_{\theta_k}(\hat p_i),y_i\big)}_{\text{NLL}}
+
\underbrace{\lambda\lVert \theta_k-\theta_0\rVert_2^2}_{\text{Shrinkage}}
\end{equation}
with \(\lambda\ge 0\) tuned on \(\mathcal D_{\text{cal}}\). This hierarchical shrinkage stabilizes estimates for small/imbalanced clusters while allowing meaningful departures from the global mapping where cluster-specific bias is present. Optimization uses L-BFGS with analytic gradients. 

\subsection{Inference (mixture of calibrators)}
At test time, for a new \(x\) we compute \(\hat p=f(x)\), the representation \(\phi(x)\), and soft memberships \(w_k(x)\). The final calibrated probability is the mixture
\[
p^{\star}(x) = \sum_{k=1}^K w_k(x) c_{\theta_k}\big(\hat p(x)\big).
\]
Algorithm \ref{alg:cluster-calib} summarizes the full Clustered Calibration procedure.

\subsection{Population-level justification}
At the population level, our use of cluster information is supported by a simple information-theoretic argument formalized in Theorem~\ref{thm:cluster-dominates-global}
in Appendix~\ref{app:theory-cluster-calib}. Let $S$ denote the base model score and let $R$ denote the cluster information derived from the learned representation $\phi(X)$. Under the Brier score, the population-optimal global calibrator is
$g_0^\star(S) = \mathbb{E}[Y \mid S]$, while the population-optimal
cluster-aware calibrator is $g_1^\star(S,R) = \mathbb{E}[Y \mid S,R]$.
Because $(S,R)$ contains at least as much information as $S$ alone, the theorem shows that
\[
\mathbb{E}\bigl[(Y - g_1^\star(S,R))^2\bigr]
\le
\mathbb{E}\bigl[(Y - g_0^\star(S))^2\bigr],
\]
i.e., cluster-aware calibrator is at least as close to Bayes as a global calibrator.
In practice, we only observe a finite calibration set and we restrict attention to parametric calibrators. The cluster-wise loss $\mathcal{L}_k(\theta_k)$ in Eq.~(\ref{eq:CCL-Loss}) is therefore a finite-sample surrogate for fitting $\mathbb{E}[Y \mid S,R]$ within the chosen parametric family, while the $\ell_2$ penalty $\lambda\|\theta_k - \theta_0\|_2^2$ acts as a
variance-reduction regularizer that shrinks small or noisy clusters toward the global calibrator $c_{\theta_0}$.

\subsection{Computational considerations}
\begin{itemize}
    \item Coverage pipeline: extraction is \(O(T)\) per sample (one leaf per tree). The CSR matrix has exactly \(T\) non-zeros per row. TF-IDF is linear in the number of non-zeros. TruncatedSVD with \(r\) components costs \(O(\mathrm{nnz}\cdot r)\).
    \item for an XGBClassifier (tree ensemble with $T$ trees, max depth $D$, and up to $L$ leaves per tree), exact TreeSHAP \cite{NIPS2017_7062} runs in  $O(TLD^2)$ time and uses about $O(D^2+M)$ memory ($M$ = \#features). For $N$ instances, multiply by $N$.
    \item Clustering: K-Means over \(Z\in\mathbb{R}^{n\times r}\) is \(O(nKr)\).
    \item Calibration: closed-form gradients yield fast L-BFGS for each \(\theta_k\). The total cost scales with \(\sum_k n_k\) where \(n_k=\sum_i w_k(x_i)\).
    \item The inference cost scales as $O(K \cdot L)$ per sample, independent of calibration set size, since only cluster centroids and calibrator parameters are stored.
\end{itemize}

\begin{algorithm}[ht]
\caption{Clustered Calibration with Shrinkage and Soft Memberships}
\label{alg:cluster-calib}
\begin{algorithmic}[1]
\Require Trained classifier $f$, calibration set $\mathcal{D}_{\mathrm{cal}}$, representation $\phi$, number of clusters $K$, shrinkage $\lambda$, calibrator family $c_\theta$
\State \textbf{(Preprocessing)}
\State $Z_{\mathrm{cal}} \gets$ embedding from $\phi$ on $\mathcal{D}_{\mathrm{cal}}$ \Comment{TF-IDF/$\ell_2$, SVD}
\State $\{\boldsymbol{\mu}_k\}_{k=1}^K \gets \mathrm{KMeans.fit}(Z_{\mathrm{cal}})$
\State Fit global calibrator $\theta_0$ on $\mathcal{D}_{\mathrm{cal}}$ by minimizing $\mathrm{NLL}\big(c_{\theta_0}(f(x)), y\big)$

\State \textbf{(Cluster-wise calibrators with shrinkage)}
\For{$k=1$ to $K$}
  \State Compute soft weights $w_k(x)$ on $\mathcal{D}_{\mathrm{cal}}$ using distances to $\boldsymbol{\mu}_k$
  \State $\theta_k \gets \arg\min_{\theta}\ \sum_{(x,y)\in\mathcal{D}_{\mathrm{cal}}} w_k(x)\,\mathrm{NLL}\!\big(c_{\theta}(f(x)), y\big) + \lambda\|\theta - \theta_0\|_2^2$
\EndFor

\State \textbf{(Inference)}
\For{each test $x$}
  \State $Z \gets$ normalized embedding $\phi(x)$. Compute $w_k(x)$ using $\{\boldsymbol{\mu}_k\}$
  \State $p^\star(x) \gets \sum_{k=1}^K w_k(x)\, c_{\theta_k}\!\big(f(x)\big)$
\EndFor
\Ensure Calibrated probabilities $p^\star(x)$
\end{algorithmic}
\end{algorithm}

\subsection{Summary}
Our method uses data-driven subpopulation discovery (via learned representations) to train a soft ensemble of calibrated experts regularized by a hierarchical prior toward a global mapping. The combination targets heterogeneous mis-calibration while remaining lightweight and reproducible. 

\section{Evaluation Metrics and Proper Scoring Rules}
\label{sec:ECE-mis-rank}
\subsection{Evaluation}
We use negative log-loss as the primary calibration metric rather than ECE variants. Unlike ECE, log-loss (NLL) is independent of any partitioning scheme, enabling fair comparison across methods with different calibration strategies (see \ref{subsubsection:ece_limitations}).  By evaluating Brier and NLL scores, we explicitly capture improvements in calibration quality that are most meaningful to decision-makers, those that enhance the accuracy of individual probability estimates within distinct subpopulations. Moreover, proper scores avoid pitfalls associated with binning-based metrics, such as sensitivity to bin sizes or arbitrary boundaries, thus providing a stable and differentiable measure of calibration performance.

\subsubsection{When ECE Mis-Ranks Soft Clustered Calibrators}
Consider two latent regions \(A,B\) with prevalence \(p_A=0.7\), \(p_B=0.3\) and equal mass. A global, uncalibrated predictor outputs (0.5) for all samples, attaining \(\text{NLL}=0.693\) and \(\text{ECE}=0\) under equal-width probability bins because its predicted probability equals the overall prevalence, so bin accuracy equals bin confidence. A soft cluster-aware calibrator predicts \(0.6\) on \(A\) and \(0.4\) on \(B\), reducing NLL to \(0.632\) but increasing ECE to \(0.1\) because fixed global bins penalize the symmetric under/over-shoot \((|0.7-0.6|=|0.3-0.4|=0.1)\). This illustrates that proper scores (e.g., NLL) can reward locally improved probabilities while ECE, being bin-based and non-proper, can move in the opposite direction when calibration is driven by soft neighborhood structure and shrinkage.
\\\\
\noindent A formal statement and a proof sketch appear below. Full proof in App. \ref{app:ece_misrank_full}.

\paragraph{Setup}
\begin{itemize}
    \item Data: Two equally sized latent regions $A,B$. True positive rates: \(p_A=0.7, p_B=0.3\).
    \item Before calibration (Global): predict a single probability for everyone, \[
 \hat {p}_{\text{before}}(x)=0.5 \quad \forall x.
 \]
    \item After calibration (Local soft + shrinkage): predict closer to the region truths, but not perfectly (“soft memberships + shrinkage to global”),
\[
\hat p_{\text{after}}(x)= \begin{cases}
 0.6 & x \in A \\
 0.4 & x \in B
 \end{cases}
\]

This mirrors a realistic outcome of the method when \(\lambda>0\): predictions move toward 0.7/0.3 but don’t hit them exactly.

\end{itemize}

\paragraph{What the metrics say}

\subparagraph{Negative log-likelihood (proper score)}

\begin{itemize}
    \item Before: \( \text{NLL} = -\big[0.5\log 0.5 + 0.5\log(1-0.5)\big] = \log 2 \approx \mathbf{0.6931}\).
    \item After (same value in A and B by symmetry):
\[
 \text{NLL} = -\big[0.7\log 0.6 + 0.3\log 0.4\big] = 0.7\cdot 0.5108 + 0.3\cdot 0.9163 
 \approx \mathbf{0.6325}.
\]
Result: NLL decreases (improves) from 0.6931 → 0.6325.
\end{itemize}

\subparagraph{ECE with fixed, equal-width probability bins (e.g., 5 bins)}

\begin{itemize}
    \item Before: all predictions fall in the central bin around 0.5. The overall empirical accuracy is 0.5 (since the mixture is 50\% of 0.7 and 50\% of 0.3).

Bin conf = 0.5, bin acc = 0.5 $\Rightarrow$ ECE = 0.

 ECE(before) = \textbf{0}.
    \item After: predictions occupy two bins, \~0.4 and \~0.6.

In the \~0.4 bin (region (B)): conf = 0.4, acc $\approx$ 0.3 $\Rightarrow$ |acc-conf| = 0.1, weight $\approx$ 0.5.

In the \~0.6 bin (region (A)): conf = 0.6, acc $\approx$ 0.7 $\Rightarrow$ |acc-conf| = 0.1, weight $\approx$ 0.5.

ECE(after) $\approx 0.5 \cdot 0.1 + 0.5 \cdot 0.1 = \mathbf{0.1}$.
\end{itemize}
Result: ECE increases from \~0 to \~0.1 after calibration.

\paragraph{Why this happens}
The after-calibration predictor is closer to the truth per region (good for NLL), but it under-shoots in \(A\) (0.6 vs 0.7) and over-shoots in \(B\) (0.4 vs 0.3). Fixed global probability bins ignore the soft, region-level conditioning that the calibrator used. They just average by predicted probability ranges. That binning sees symmetric \(\pm0.1\) deviations and penalizes them, even though the per-sample likelihood improved (see table \ref{tab:NLL-ECE-Toy Demo}).
\begin{table}[ht]
\centering
\begin{tabular}{ccc}\toprule
Model & NLL & ECE (5 bins) \\\midrule
Before (Global) & 0.6931 & 0.0000 \\
After (Local + $\lambda > 0$) & 0.6325 & 0.1000 \\ \bottomrule
\end{tabular}
\caption{Comparison of NLL vs. ECE before and after soft clustered calibration.}
\label{tab:NLL-ECE-Toy Demo}
\end{table}

\paragraph{Practical takeaway}
For soft, representation-aware calibrators such as clustered calibration, we recommend using proper scoring rules (negative log-likelihood or Brier score) as the primary metrics for model selection and reporting, and using ECE mainly as a diagnostic visualization (via reliability diagrams) rather than as a stand-alone optimization target. Our toy example and large-scale experiments both show that fixed-bin ECE can increase even when the underlying probabilistic predictions become closer to the true conditional distribution.

\section{Experiments and Results}
\subsection{Data and code availability}
\subsubsection{Data availability }
All datasets used in this study are publicly available (See table \ref{tab:datasets}. Clinical datasets were used under data use agreements with their providers. Where applicable, derived feature matrices and scripts to reproduce the experiments are available in the accompanying code repository, subject to any institutional data-sharing restrictions.

\subsubsection{Code availability}
The implementation of clustered calibration, together with experiment scripts and configuration files to reproduce all results in this paper, is available at: \url{https://anonymous.4open.science/r/clustered-calibration-EE9C/}.

\subsubsection{Experiments results}
The result files of the experiments performed in this paper are available in the accompanying code repository.

\subsection{Experimental setup}
To evaluate our method performance and robustness we performed a wide series of experiments comparing our method against strong baseline methods on a suite of benchmarks (See table \ref{tab:datasets}) and a range of model architectures. 
Our primary empirical focus is on tabular classification, where we observe the most pronounced and consistent gains in NLL and Brier score across model capacities. The image and text experiments play a complementary role: they probe how clustered calibration behaves in deep representation spaces and whether the same framework remains competitive when the learned representations and label structures differ substantially from tabular settings.

\begin{itemize}
    \item Tasks: binary and multi-class classification.
    \item Models: 
    \begin{itemize}
        \item eXtreme Gradient Boosting (XGBoost) \cite{chen2016xgboost} with fixed grids:  n\_estimators (100/300/1000), max\_depth (6/8), learning\_rate (0.1), subsample=0.8, colsample\_bytree=0.8.
        \item ResNet-50 \cite{he2016deep} pretrained on ImageNet and fine-tuned using 0.0001 learning rate, for up to 100 epochs, using early stopping if validation performance does not improve for 5 epochs.
        \item DistilBERT \cite{sanh2019distilbert} fine-tuned using 0.0002 learning rate, 0.01 weight decay and 500 warmup steps for 3 epochs.
    \end{itemize}
    \item Splits: five repeated holdouts with fixed seeds:
    \begin{itemize}
        \item For tabular datasets: 60\% for training, 20\% for calibration and 20\% as a held-out test set.
        \item For fine-tuned models: validation and test sets were merged and re-split to calibration and test set of size 50\% each.
    \end{itemize}
    \item For each combination of dataset, method and model configuration, we aggregate the metrics by taking the mean, standard deviation and a 95\% t-based confidence interval over the repeated random seeds within that group.
    \item All reported result values are means with 95\% CI over datasets and model architectures.
    \item Tabular data preprocessing:
    \begin{itemize}
        \item Fill missing values using most frequent values.
        \item Encode categorical features.
        \item All features are standardized.
    \end{itemize}
    \begin{itemize}
    \item Dataset-specific preprocessing:
    \begin{itemize}
        \item LOS: binary label was computed by length of stay over 5 days.
    \end{itemize}
    \end{itemize}
    \item Clustering:
    \begin{itemize}
        \item Clustering data (see Section \ref{sec:method}):
        \begin{itemize}
            \item Tabular: coverage and SHAP vectors as embedded values, as well as raw data points as an ablation test.
            \item Image and Text: pre-classification head activation and embedding vectors.
        \end{itemize}
        \item Clustering method: K-means ~\cite{lloyd1982least} for a default of $K$=4 partitions, as well as additional $K$ values as a hyperparameter optimization and sensitivity tests.
    \end{itemize}
    \item Calibration: we applied our soft mixture of per-cluster calibrators with hierarchical shrinkage regularization, while omitting the shrinkage regularization as an ablation test.
    \item Baseline calibration methods - A total of 21 calibration methods were experimented: Histogram Binning, Bayesian binning, Isotonic Calibration, Ensemble TS. BTS and our Clustered Calibration (coverage, SHAP, data, activations and hidden states) were implemented, each using Platt/Vector, TS and Beta/Dirichlet as base methods. 
    \item Calibration baselines all ran with default hyperparameters:
    \begin{itemize}
        \item Histogram Binning: Posterior mean of $Beta(\alpha,\beta)$ prior with Bernoulli likelihood with defaults $\alpha=1, \beta=1$.
        \item Bayesian Binning: score\_function='BIC', equal\_intervals=True, detection=False, independent\_probabilities=False.
        \item Binwise methods: n\_bins=10, min\_bin\_size=30.
        \item Dirichlet: weigh\_decay=1e-3.
        \item IsotonicRegression: out\_of\_bounds='clip'.
        \item Ensemble TS: loss='ce', bounds=(0.05, 5.0).
        \item Beta: parameters='abm'.
    \end{itemize}
    \item Compute: All experiments were performed on a MacbookPro M4 Max 36GB RAM. 
\end{itemize}

\subsubsection{Dirichlet calibration implementation}
We used the authors' reference code as well as implementing Dirichlet calibration following \cite{kull2019beyond}.\footnote{We used the publicly released implementation from \cite{kull2019beyond} and our own re-implementation, obtaining qualitatively similar behavior and in some cases slightly worse performance with the reference code.} 
Across our experiments, both implementations produced highly consistent results, suggesting that the observed behavior of Dirichlet calibration, especially on high class-count benchmarks such as CIFAR-100 and ImageNet, is not an artifact of our implementation but reflects genuine sensitivity of the method in these regimes. 

\begin{table}[ht]
  \centering

\begin{tabular}{lccc}
        \toprule
        \textbf{Name} & \textbf{Domain} & \textbf{Features} & \textbf{Samples}\\
        \midrule
         \multicolumn{4}{c}{\textbf{Tabular}}\\
        \midrule
        Adult \cite{adult_2} & Social & 14 & 48,842 \\
        Credit Default \cite{default_of_credit_card_clients_350} & Business & 15& 690\\
        Diabetes 130 \cite{diabetes_130-us_hospitals_for_years_1999-2008_296} & Health & 47 & 101,766 \\
        ICU stays (LOS) \cite{mimiciv_v04} & Health & 14 & 45,889 \\
        Stroke \cite{mxfb} & Health & 12 & 5,110 \\
        WiDS Datathon \cite{widsdatathon2020} & Health & 186 & 91,713 \\
        \midrule
         \multicolumn{4}{c}{\textbf{Images}}\\
        \midrule
        BloodMNIST \cite{bloodmnist} & Health & N/A& 60,000 \\
        CIFAR-100 \cite{krizhevsky2009learning} & Misc & N/A& 17,092 \\
        ImageNet \cite{ImageNet} & Misc & N/A& $\approx$1.2M \\
        \midrule
         \multicolumn{4}{c}{\textbf{Text}}\\
        \midrule
        Emotion \cite{saravia-etal-2018-carer} & Emotions & N/A& 19,930 \\
        IMDB \cite{imdb} & Sentiment & N/A& 50,000 \\
        \bottomrule
    \end{tabular}
    \caption{Overview of datasets with various domains.}
    \label{tab:datasets}
\end{table}

\subsection{Results}\label{experimental-results}
We first present results on tabular data, which constitute our primary empirical case study, and then report complementary experiments on image and text benchmarks.

\subsubsection{Tabular classification results}
To quantify the effect of clustered calibration, we aggregated results across 6 datasets, 6 hyperparameter configurations, and 5 repeated experiments. Aggregating results across all 5,780 data points (6 datasets × 21 methods × 6 configurations × 5 repeats) reveals consistent superiority of clustering-based calibration methods.

Table \ref{tab:performance} presents the overall performance ranking by log-loss, the primary evaluation metric emphasizing calibration quality. Clustered methods dominate the top rankings, occupying top positions overall, demonstrating systematic superiority over baseline approaches. Lower is better for log-loss, Brier and ECE. Higher is better for AUC and Accuracy.

Averaged over the 6 datasets, the uncalibrated base model attains a mean log-loss of 0.376, whereas standard global calibrators such as Platt Scaling and Beta calibration reduce log-loss to $\approx 0.300$, i.e., a $\approx 20\%$ improvement, while preserving AUC around 0.82 and substantially lowering Brier. Cluster-aware variants then provide a systematic gain on top of these strong baselines: clustered Platt and Beta calibration using SHAP or coverage representations achieve the lowest mean log-loss values ($\approx 0.298-0.299$), corresponding to $\approx 0.3-0.5\%$ additional improvement over the best global calibrators, with equal or slightly higher AUC and virtually unchanged Brier scores. Clustered Temperature Scaling exhibits a similar pattern, yielding $\approx 0.4-1.3\%$ reductions in log-loss relative to global Temperature Scaling. Overall, this analysis shows that log-loss is sensitive enough to detect the consistent benefits of cluster-aware calibration, and that these gains are obtained without sacrificing discrimination performance.

\begin{table}[ht]
\centering

\begin{tabular}{c c c c c c c}\toprule
\textbf{Method} & \textbf{Clustered} & \textbf{NLL}$\downarrow$ & \textbf{AUC}$\uparrow$ & \textbf{Brier}$\downarrow$ & \textbf{ECE}$\downarrow$& \textbf{LCE}$\downarrow$ \\\midrule
Platt & SHAP & \textbf{0.2983} & \textbf{0.8244} & \textbf{0.0897} & 0.0266 & 0.6620 \\
Platt & Coverage & \textbf{0.2983} & 0.8232 & \textbf{0.0897} & 0.0257 & 0.5725 \\
Beta & Coverage & 0.2986 & 0.8231 & 0.0898 & 0.0259 & 0.5710 \\
Platt & Data & 0.2988 & 0.8215 & \textbf{0.0897} & 0.0261 & 0.7318 \\
Beta & Data & 0.2989 & 0.8217 & 0.0898 & 0.0271 & 0.7385 \\
Beta & SHAP & 0.2990 & 0.8231 & 0.0900 & 0.0276 & 0.6672 \\
Beta & Baseline & 0.2997 & 0.8208 & 0.0898 & 0.0259 & 0.5776 \\
Platt & Baseline & 0.2998 & 0.8207 & 0.0898 & \textbf{0.0256} & 0.5770 \\
Histogram Binning & Baseline & 0.3044 & 0.7901 & 0.0909 & 0.0274 & \textbf{0.5381} \\
Binwise-Beta & Baseline & 0.3074 & 0.8089 & 0.0919 & 0.0294 & 0.5692 \\
Binwise-Platt & Baseline & 0.3076 & 0.8081 & 0.0920 & 0.0295 & 0.5680 \\
Bayesian Binning & Baseline & 0.3109 & 0.7856 & 0.0919 & 0.0291 & 0.5593 \\
Binwise-TS & Baseline & 0.3240 & 0.8097 & 0.0995 & 0.0828 & 0.5907 \\
LoRe & Baseline & 0.3256 & 0.8099 & 0.0946 & 0.0357 & 0.5873 \\
TS & SHAP & 0.3410 & 0.8221 & 0.1066 & 0.1191 & 0.6856 \\
Isotonic & Baseline & 0.3422 & 0.8175 & 0.0909 & 0.0283 & 0.5654 \\
TS & Coverage & 0.3425 & 0.8219 & 0.1071 & 0.1208 & 0.6553 \\
TS & Data & 0.3441 & 0.8211 & 0.1075 & 0.1221 & 0.7542 \\
ensemble-TS & Baseline & 0.3447 & 0.8208 & 0.1076 & 0.1223 & 0.6543 \\
TS & Baseline & 0.3456 & 0.8208 & 0.1077 & 0.1221 & 0.6543 \\
base & Baseline & 0.3756 & 0.8208 & 0.1104 & 0.1303 & 0.6736 \\ 
\bottomrule

\end{tabular}
\caption{Overall Performance. Clustering methods ranked by mean log-loss.}
\label{tab:performance}
\end{table}

Table \ref{tab:configurations} shows that the best clustered method for each capacity consistently improves log-loss by about 0.5-1.0\% relative to the best non-clustered baseline, while also yielding consistent AUC gains. With the largest improvement observed in the n\_estimators=300/1000, max\_depth=8 configuration, the results demonstrate that clustering benefits increase with model complexity.

Beyond log-loss, clustered methods show even stronger performance on AUC, the discriminative performance metric. Clustered methods achieve best AUC in all 6 configurations. The AUC improvements show a clear trend toward larger gains in more complex models: improvements of +0.67\% to +0.95\% are observed for n\_estimators=1000, compared to +0.06\% to +0.15\% for n\_estimators=100. 

Consistent with log-loss and AUC improvements, clustering also maintains or moderately improves Brier score. This consistency across all three metrics indicates that clustering provides genuine calibration improvement rather than metric-specific artifacts.
The inconsistency of the ECE comparison aligns with our claim that ECE is not the primary measure for soft cluster assignment methods.

\begin{table}[t]
\centering

\begin{tabular}{cc r r llr r r r }
\toprule
$n$ & $d$ &
$\text{NLL}_{\text{non}}$ &
$\text{NLL}_{\text{cl}}$  & $\text{Brier}_{\text{non}}$ &$\text{Brier}_{\text{cl}}$  &
$\text{AUC}_{\text{non}}$ &
$\text{AUC}_{\text{cl}}$ &
$\text{ECE}_{\text{non}}$ &
$\text{ECE}_{\text{cl}}$ \\
\midrule
100  & 6 &
0.2940 & \textbf{0.2926}  & 0.0882 &\textbf{0.0881}  &
0.8314 & \textbf{0.8318} &
0.0281 & \textbf{0.0268} \\
100  & 8 &
0.2945 & \textbf{0.2928}  & 0.0884 &\textbf{0.0883}  &
0.8289 & \textbf{0.8302} &
0.0258 & \textbf{0.0246} \\
300  & 6 &
0.2982 & \textbf{0.2965}  & \textbf{0.0890} &\textbf{0.0890}  &
0.8228 & \textbf{0.8249} &
\textbf{0.0246} & 0.0265 \\
300  & 8 &
0.3007 & \textbf{0.2978}  & 0.0899 &\textbf{0.0895}  &
0.8189 & \textbf{0.8225} &
0.0253 & \textbf{0.0234} \\
1000 & 6 &
0.3043 & \textbf{0.3029}  & \textbf{0.0911} &\textbf{0.0911}  &
0.8122 & \textbf{0.8176} &
\textbf{0.0238} & 0.0280 \\
1000 & 8 &
0.3064 & \textbf{0.3034}  & 0.0919 &\textbf{0.0915}  &
0.8109 & \textbf{0.8186} &
\textbf{0.0261} & 0.0265 \\
\bottomrule
\end{tabular}
\caption{Best clustered vs.\ best non-clustered calibration per model capacity. $NLL_{non}/NLL_{cl}$ denote the best non-clustered and clustered log-loss, respectively. Similarly for Brier and ECE}
\label{tab:configurations}
\end{table}

\paragraph{Statistical Significance Tests}
For each tabular dataset and model capacity (n\_estimators $\in$ \{100, 300, 1000\}, max\_depth $\in$ \{6, 8\}), we selected the best global parametric calibrator (Platt, Beta or Temperature Scaling) and the best clustered variant (same base methods with SHAP/coverage/data representations), based on validation log-loss. Across all 6 datasets × 6 capacities (36 paired configurations), clustered calibration strictly reduced negative log-loss and improved AUC in every case (36/36 wins for both metrics), with a median absolute NLL decrease of $\approx 6\times10^{-4}$ ($\approx0.24\%$ relative) and a median absolute AUC gain of $\approx 8.5\times10^{-4}$ ($\approx0.10\%$ relative). A Wilcoxon signed-rank test on these 36 pairs confirms that the NLL and AUC improvements are highly significant ($p \approx 2.9\times10^{-11}$ for both), and remain significant when averaging over capacities per dataset ($p = 0.031$ over 6 datasets). Brier score also improves in 34/36 configurations (Wilcoxon $p \approx 6.3\times10^{-7}$), while ECE changes are mixed and not statistically significant ($p \approx 0.62$), in line with our critique that bin-based ECE does not reliably reflect the gains of soft, representation-based clustered calibration.

Table \ref{tab:tabular_clustered_vs_global} summarizes the relative improvements of clustered calibration over the best global parametric calibrator on the six tabular datasets. Across datasets, clustered calibration yields consistently positive gains in the proper scoring rules: averaged over all 6 datasets and 6 capacities, log-loss improves by $0.49\%$ (95\% CI [0.29, 0.68]) and Brier score by $0.12\%$ (95\% CI [0.05, 0.19]), while AUC increases by $0.43\%$ (95\% CI [0.19, 0.68]), with the largest gains observed on Stroke and Diabetes130. In contrast, ECE exhibits high variability and is on average worse under clustered calibration ($-4.56\%$, 95\% CI [-8.55, -0.57]), reinforcing our claim that ECE does not reliably track improvements in proper scoring rules when calibration is performed via soft cluster assignments.

\begin{table}[t]
\centering

\begin{tabular}{lcccc}
\toprule
Dataset & $\Delta$NLL (\%) & $\Delta$AUC (\%) & $\Delta$Brier (\%) & $\Delta$ECE (\%) \\
\midrule
Adult           & $0.12 \pm 0.11$ & $0.02 \pm 0.03$ & $0.12 \pm 0.12$  & $1.19 \pm 3.06$ \\
Credit          & $1.55 \pm 0.28$ & $0.06 \pm 0.06$ & $-0.09 \pm 0.26$ & $-0.72 \pm 5.90$ \\
Diabetes130     & $0.17 \pm 0.09$ & $0.93 \pm 0.63$ & $0.11 \pm 0.07$  & $-9.15 \pm 12.24$ \\
LOS             & $0.16 \pm 0.19$ & $0.12 \pm 0.15$ & $0.17 \pm 0.20$  & $-3.49 \pm 7.81$ \\
Stroke          & $0.76 \pm 0.48$ & $1.38 \pm 1.21$ & $0.34 \pm 0.25$  & $-3.58 \pm 9.55$ \\
WiDS            & $0.16 \pm 0.12$ & $0.10 \pm 0.09$ & $0.09 \pm 0.08$  & $-11.59 \pm 23.80$ \\
\midrule
All (6 tabular) & $0.49 \pm 0.19$ & $0.43 \pm 0.25$ & $0.12 \pm 0.07$  & $-4.56 \pm 3.99$ \\
\bottomrule
\end{tabular}
\caption{Relative improvement (\%) of clustered calibration over the best global parametric calibrator on the six tabular datasets. For each dataset we report the mean percentage change across capacities together with a 95\% confidence interval. Positive values indicate that clustered calibration improves the metric, while negative values for ECE indicate that it tends to worsen under clustered calibration.}
\label{tab:tabular_clustered_vs_global}
\end{table}

\paragraph{Practical Example: CDSS}
\label{example:CDSS}
On the Stroke dataset, clustering the learned representations yields four clinically distinct subpopulations (Table \ref{tab:stroke-clusters}). Cluster 0 consists mainly of older patients (60-78 years) with moderately elevated BMI and markedly higher glucose levels. It exhibits the highest stroke prevalence (14.9\%). Cluster 1 captures very young patients (13-29 years) with lower BMI and normal glucose ranges and contains almost no stroke events. Cluster 2 corresponds to middle-aged patients (50-62 years) with somewhat higher BMI and mildly elevated glucose and shows an intermediate stroke rate (5.5\%), while Cluster 3 includes early middle-aged patients (39-44 years) with near-normal glucose and a very low stroke rate (0.7\%). 
These cluster-specific feature ranges and outcome frequencies make the heterogeneity in risk profile explicit, providing intuition for why a single global calibration function must compromise across groups, whereas cluster-based calibration can adapt to the distinct calibration needs of each sub-population and, at the same time, expose clinically meaningful strata that may support decision making in a clinical decision support system (CDSS). From an explainability perspective, assigning a patient to one of these typical sub-groups according to their age, BMI and glucose values provides a direct, feature-based justification of the calibrated risk estimate. 

\begin{figure}[t]
\centering
\includegraphics[width=0.48\linewidth]{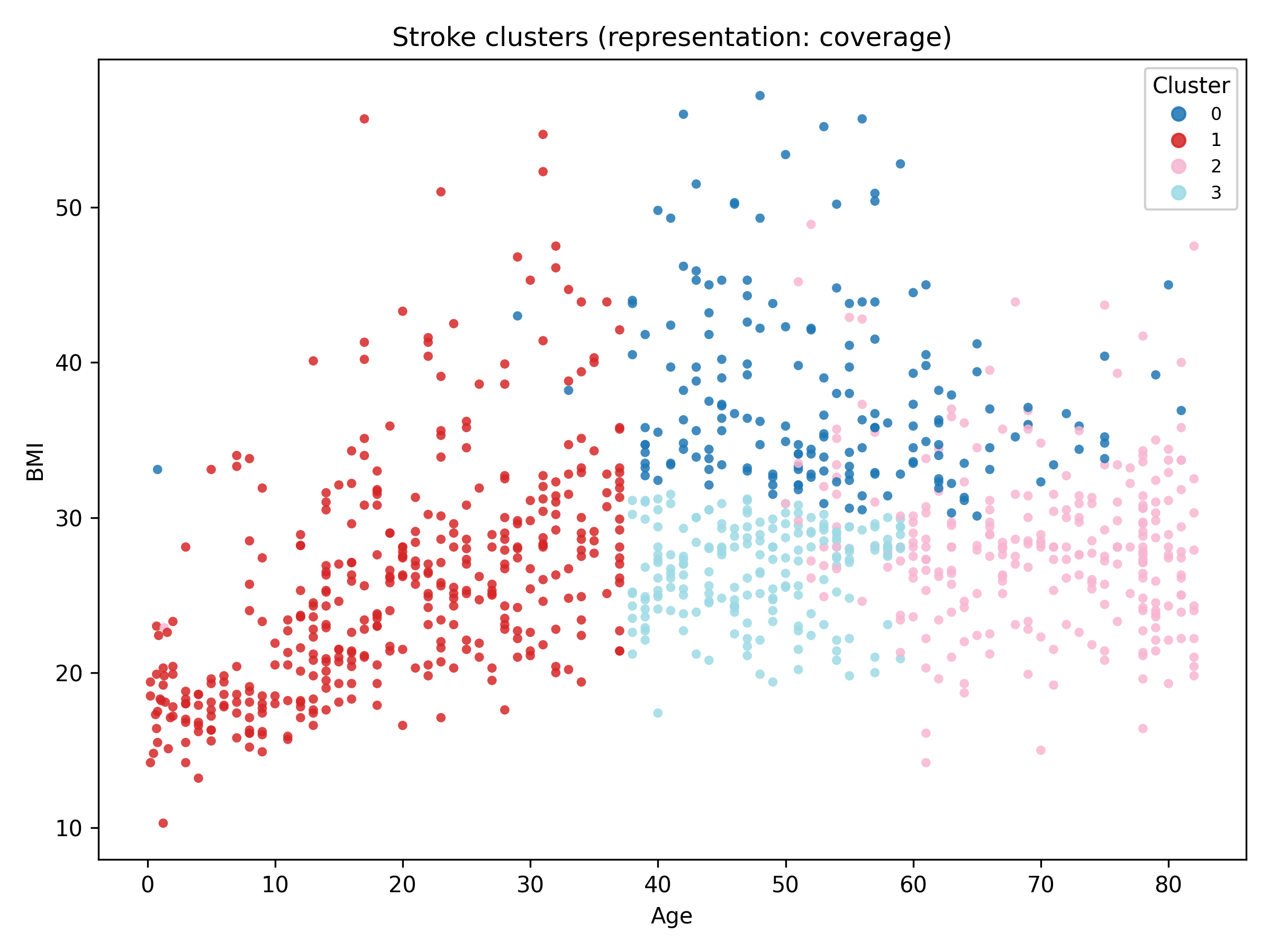}
\hfill
\includegraphics[width=0.48\linewidth]{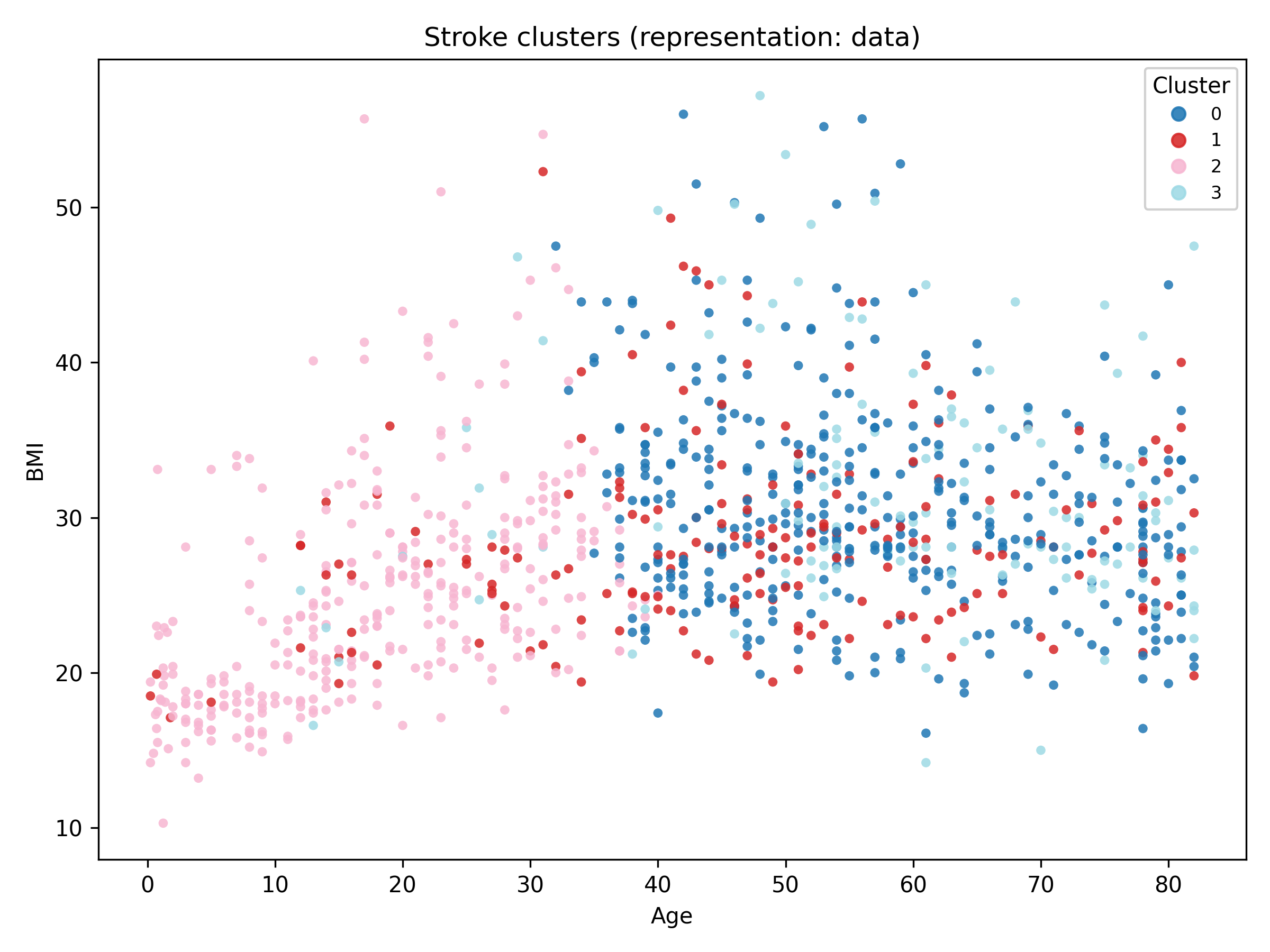}
\caption{Stroke clusters in the age-BMI plane for the coverage-based representation (left) and for clustering performed directly in the raw data space (right). The coverage-based representation yields more clearly separated and clinically coherent subpopulations.}
\label{fig:stroke-clusters-age-bmi}
\end{figure}

\begin{table}[t]
\centering

\begin{tabular}{lcccc}
\toprule
Cluster & Age (years) & BMI & Avg. glucose (mg/dL) & Stroke rate (\%) \\
\midrule
0 & 60-78 & 26.4-32.4 & 83.6-195.2 & 14.9 \\
1 & 13-29 & 20.5-29.6 & 77.5-105.7 & \phantom{0}0.0 \\
2 & 50-62 & 26.3-34.4 & 75.1-107.5 & \phantom{0}5.5 \\
3 & 39-44 & 24.3-33.5 & 77.7-109.3 & \phantom{0}0.7 \\
\bottomrule
\end{tabular}
\caption{Typical feature ranges and stroke prevalence per cluster on the Stroke dataset.
Age, BMI and glucose are reported as inter-quartile ranges (25th-75th percentile).
Stroke rate is the mean fraction of positive labels in the cluster.}
\label{tab:stroke-clusters}
\end{table}
Figure \ref{fig:stroke-clusters-age-bmi} visualizes the four clusters in the age-BMI plane for both coverage-based and raw-data representations, highlighting how the coverage-based clustering yields more distinct, clinically coherent regions and further clarifies why learned representations cluster-based calibration can better adapt to the heterogeneous calibration needs of these subpopulations than a single global calibrator.

\begin{figure}[t]
\centering
\begin{minipage}{0.32\linewidth}
  \centering
  \includegraphics[width=\linewidth]{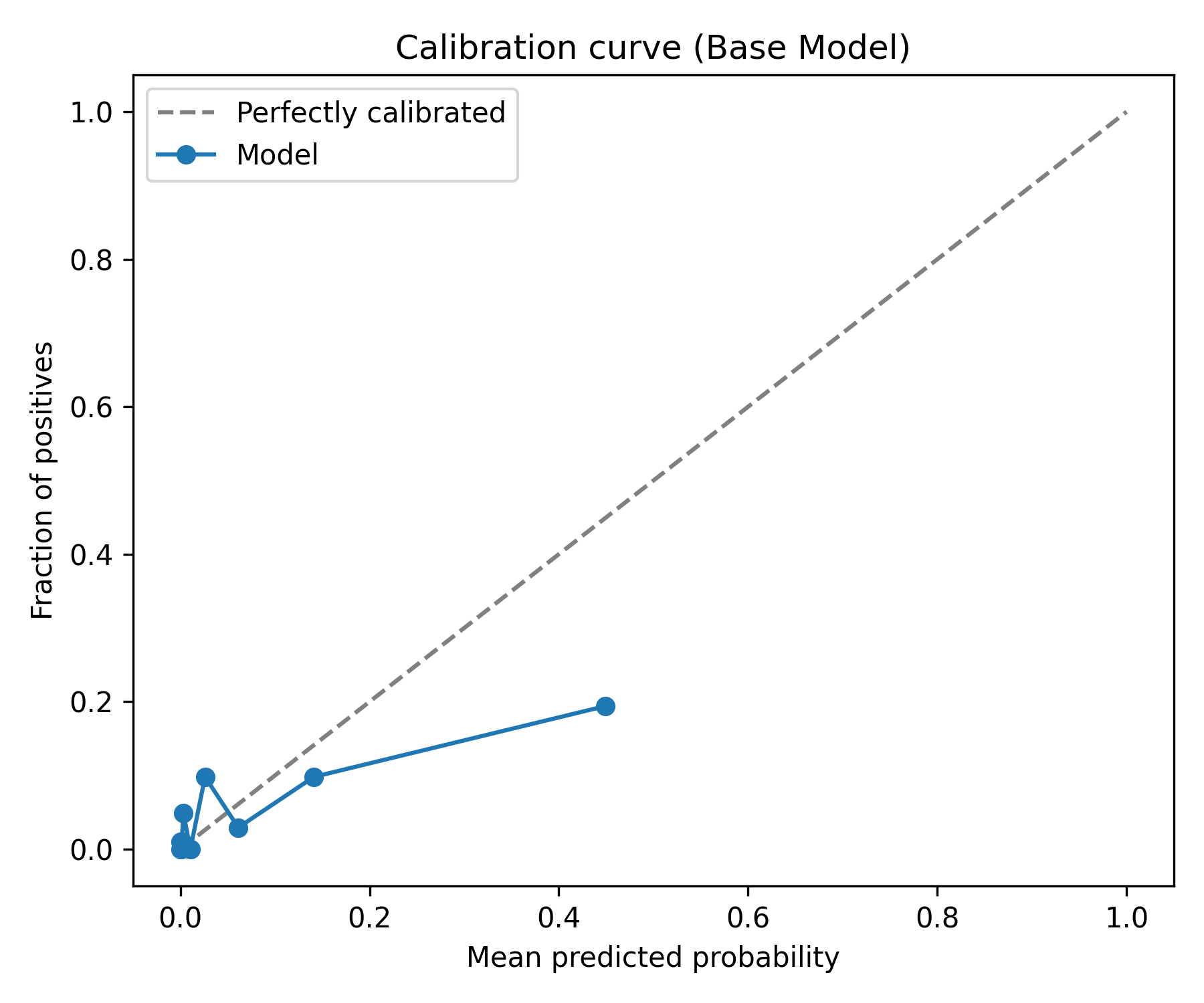}
\end{minipage}
\hfill
\begin{minipage}{0.32\linewidth}
  \centering
  \includegraphics[width=\linewidth]{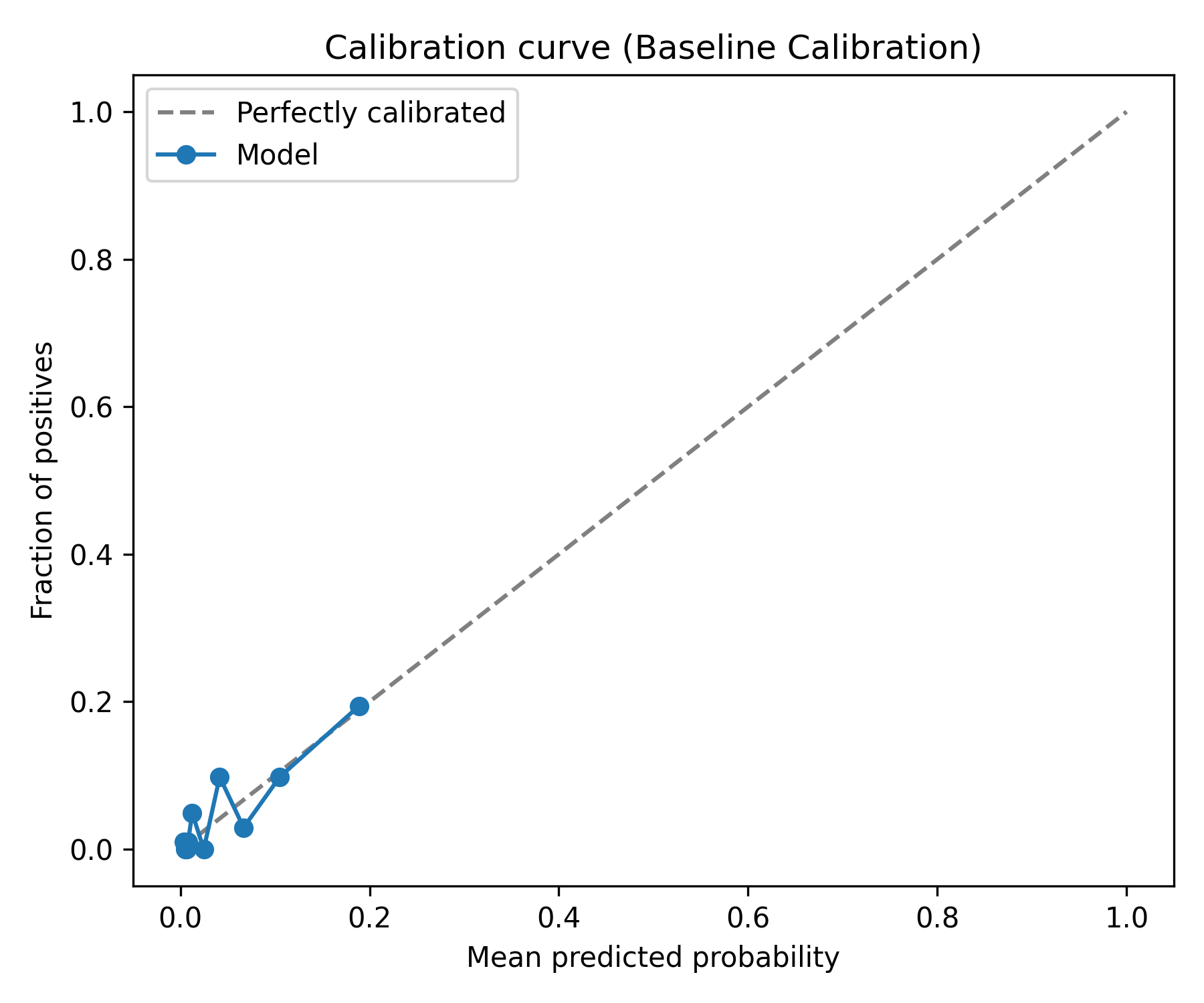}
\end{minipage}
\hfill
\begin{minipage}{0.32\linewidth}
  \centering
  \includegraphics[width=\linewidth]{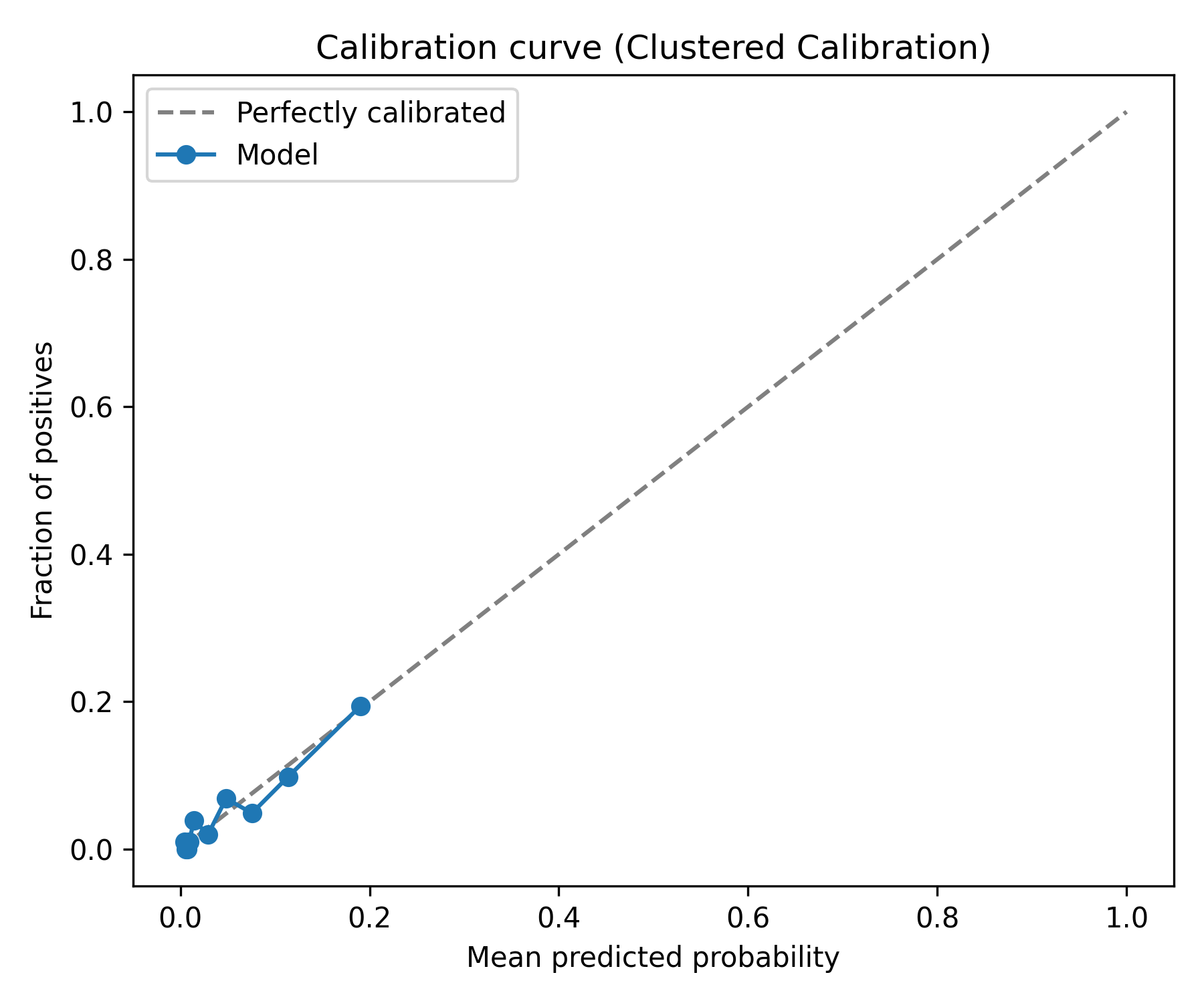}
\end{minipage}
\caption{Reliability diagrams on the Stroke dataset for the uncalibrated classifier (left),
global Beta calibration (middle), and clustered Beta calibration (right). Clustered
Beta produces curves closest to the diagonal, indicating the best alignment between
predicted stroke risk and observed frequencies.}
\label{fig:stroke-calibration-curves}
\end{figure}

Next, we demonstrate how clustered calibration contributes to CDSS on various sample rejection thresholds. Table \ref{tab:stroke-rejection} shows that Beta calibration substantially reduces the error rate compared to the uncalibrated classifier on upper rejection levels, and that clustered Beta contributes to the global variant, yielding the lowest errors for the rejection quantiles (0.3-0.5). Figure \ref{fig:stroke-calibration-curves} shows the corresponding reliability diagrams, where the clustered Beta variant lies closest to the diagonal across all bins, visually confirming that cluster-based calibration delivers the most faithfully calibrated stroke-risk probabilities.

\begin{table}[t]
\centering

\begin{tabular}{lccc}
\toprule
Rejection quantile & Base error & Beta (global) & Beta (clustered) \\
\midrule
0.0 & 0.0675 & 0.0489 & 0.0489 \\
0.1 & 0.0381 & 0.0326 & 0.0326 \\
0.2 & 0.0269 & 0.0245 & 0.0245 \\
0.3 & 0.0252 & 0.0238 & \textbf{0.0210} \\
0.4 & 0.0114 & 0.0114 & \textbf{0.0113} \\
0.5 & 0.0137 & 0.0137 & \textbf{0.0117} \\
0.6 & 0.0049 & 0.0049 & 0.0049 \\
0.7 & 0.0033 & 0.0033 & 0.0033 \\
0.8 & 0.0049 & 0.0049 & 0.0049 \\
0.9 & 0.0097 & 0.0097 & 0.0097 \\
\bottomrule
\end{tabular}
\caption{Classification error rate as a function of rejection quantile on the Stroke dataset (0 = accept all, 1 = reject all). Lower values are better.}
\label{tab:stroke-rejection}
\end{table}

\subsubsection{Deep models}
\paragraph{Image classification results}
To assess whether our conclusions extend beyond tabular data, we also evaluate clustered calibration on three image classification benchmarks: ImageNet, CIFAR-100, and BloodMNIST from the MedMNIST collection \cite{medmnistv2}. We start from strong pretrained or fine-tuned models and apply the leading set of calibration methods used in the tabular experiments: uncalibrated base predictions, bin-wise Vector Matrix and Temperature Scaling baselines, global parametric calibrators (Vector, TS, Dirichlet), and their clustered variants. Table~\ref{tab:image_calibration} summarizes negative log-likelihood (NLL), Brier score, ECE and accuracy, averaged over five random calibration-test splits.

All calibration methods substantially improve over the uncalibrated base models in terms of NLL and Brier, while largely preserving accuracy. Clustered parametric calibrators (in particular clustered Vector scaling) are consistently competitive with or slightly better than strong bin-wise baselines such as Binwise-TS. The additional benefit of clustering on top of a good global calibrator is positive: Clustered-Vector achieves the best or near-best NLL/Brier on CIFAR-100 and ImageNet. Again, ECE often ranks methods differently. These results mirror our tabular findings: once flexible, cluster-based calibration schemes are allowed, ECE is not as reliable as proper scoring rules such as NLL and Brier, which provide a more trustworthy basis for assessing and comparing calibrated predictors.

\begin{table}[ht]
\centering

\begin{tabular}{l l c c c c c c}\toprule
\textbf{Dataset} & \textbf{Method} & \textbf{K} & \textbf{$\lambda$} & \textbf{NLL $\downarrow$} & \textbf{Brier $\downarrow$} & \textbf{ECE $\downarrow$} & \textbf{Acc $\uparrow$} \\\midrule
BloodMNIST & Base & - & -& 0.3185 & 0.1593 & 0.0269 & \textbf{0.8940} \\
 & Binwise-Vector & - & -& 0.3334 & 0.1605 & 0.0177 & 0.8925 \\
 & Binwise-TS & - & -& 0.3305 & $\underline{0.1574}$ & \textbf{0.0141} & \textbf{0.8940} \\
 & Dirichlet & - & -& 0.3123 & 0.1579 & 0.0173 & 0.8922 \\
 & Dirichlet & 10 & 1 & \textbf{0.3119} & 0.1577 & 0.0173 & 0.8922 \\
 & Vector & - & -& $\underline{0.3121}$ & \textbf{0.1573} & $\underline{0.0172}$ & $\underline{0.8935}$ \\
 & Vector & 10 & 1 & $\underline{0.3121}$ & \textbf{0.1573} & $\underline{0.0172}$ & $\underline{0.8935}$ \\
 & TS & - & -& 0.3156 & 0.1579 & 0.0188 & \textbf{0.8940} \\
 & TS & 10 & 1 & 0.3152 & 0.1577 & 0.0185 & \textbf{0.8940} \\
\midrule
CIFAR-100 & Base & - & -& 2.3367 & 0.6410 & 0.2607 & 0.5990 \\
 & Binwise-Vector & - & -& 1.9328 & 0.5715 & 0.0923 & 0.5871 \\
 & Binwise-TS & - & -& 1.6487 & 0.5467 & $\underline{0.0387}$ & 0.5990 \\
 & Dirichlet & - & -& 3.3938 & 0.6823 & 0.1875 & 0.5146 \\
 & Dirichlet & 10 & 0.05 & 5.2155 & 0.6876 & 0.1315 & 0.5268 \\
 & Vector & - & -& $\underline{1.6257}$ & $\underline{0.5431}$ & 0.0532 & $\underline{0.5998}$ \\
 & Vector & 10 & 0.05 & \textbf{1.6157} & \textbf{0.5412} & 0.0481 & \textbf{0.6009} \\
 & TS & - & -& 1.6418 & 0.5462 & \textbf{0.0386} & 0.5990 \\
 & TS & 10 & 0.05 & 1.6790 & 0.5475 & 0.0415 & 0.5990 \\
\midrule
ImageNet & Base & - & -& 1.4776 & 0.4872 & 0.4008 & 0.7899 \\
 & Binwise-Vector & - & -& 1.0374 & 0.3534 & $\underline{0.0313}$ & 0.7599 \\
 & Binwise-TS & - & -& \textbf{0.8690} & 0.3017 & 0.0373 & $\underline{0.7899}$ \\
 & Dirichlet & - & -& 15.8978 & 1.6724 & 0.7541 & 0.0028 \\
 & Dirichlet & 10& 5 & 35.9447 & 1.6116 & 0.7129 & 0.0027 \\
 & Vector & - & -& 0.8784 & \textbf{0.2987} & \textbf{0.0139} & \textbf{0.7919} \\
 & Vector & 10& 5 & 0.8783 & \textbf{0.2987} & \textbf{0.0139} & \textbf{0.7919} \\
 & TS & - & -& 0.8721 & $\underline{0.3015}$ & 0.0327 & $\underline{0.7899}$ \\
 & TS & 50& 5 & $\underline{0.8720}$ & $\underline{0.3015}$ & 0.0321 & $\underline{0.7899}$ \\ \bottomrule

\end{tabular}
\caption{Calibration performance on image benchmarks. We report mean NLL, Brier score, ECE, and accuracy over five random splits. $K$ denotes the number of clusters for clustered methods. $K=\text{'-'}$ indicates a non-clustered baseline method. Lower is better for NLL, Brier and ECE. 
Each entry is averaged over five random splits. 
For each method we report its best configuration (activation and shrinkage~$\lambda$) chosen by validation log-loss.
Higher is better for accuracy. Rows are unsorted, divided by dataset. Bold indicates best score and underline indicates second best.}
\label{tab:image_calibration}
\end{table}

\subparagraph{On the Dirichlet calibration performance}
A noticeable pattern in Table~\ref{tab:image_calibration} is the weak NLL of Dirichlet calibration on CIFAR-100 and ImageNet, despite reasonable accuracy. 
We verified our implementation against the original code of \cite{kull2019beyond} and observed similar or worse behavior, indicating that this is not an implementation error but a property of the method in high-class regimes.
This observation is consistent with prior work: \cite{luo2022local} report that Dirichlet calibration tends not to generalize well when the number of classes is large and highlight its sensitivity in such settings, building on empirical findings by \cite{zhao2021calibrating}. 
A plausible explanation is that Dirichlet's parameterization, which scales with the number of classes, leads to a very high-dimensional calibration model that is sensitive to class imbalance and to limited calibration data per class, making it prone to overfitting and numerical instability when applied to datasets such as CIFAR-100 and ImageNet.
We therefore retain Dirichlet as an important baseline, as a multivariate extension of Beta calibration, but interpret its performance on these benchmarks as evidence of these known limitations rather than a failure of the calibration evaluation protocol.

\paragraph{Text classification results}
We evaluate our calibration methods on two text classification benchmarks: IMDB (binary sentiment) and Emotion (six-way emotion classification). In both cases we fine-tune a DistilBERT classifier and apply the same family of post-hoc calibrators as in the other experiments: global parametric methods (Platt/Vector scaling, Temperature Scaling, Dirichlet and Beta calibration), their representation-based clustered variants, and strong binwise baselines, reporting log-loss, Brier score, accuracy, AUC (for IMDB) and ECE averaged over five random splits (see tables \ref{tab:nlp}). On IMDB, the uncalibrated model is clearly over-confident, and all parametric calibrators substantially reduce log-loss with essentially unchanged accuracy and AUC. The best global methods (Platt and Beta) reach about a 35\% relative reduction in log-loss, while clustering further improves some of these models (e.g., clustered TS and clustered Beta outperform their global counterparts), illustrating that representation-based local structure can be beneficial. The strongest overall performance, however, is obtained by the bin-wise variants, which achieve roughly a 40\% reduction in log-loss and the lowest Brier scores. On Emotion, the base model is already reasonably calibrated, yet global Vector, Dirichlet and TS still yield modest gains in log-loss and Brier and Clustered Vector provides a small additional improvement over global Vector, whereas clustered Dirichlet and clustered TS slightly degrade performance. Again, bin-wise methods attain the best overall scores, with Binwise-Vector achieving the lowest log-loss and Brier, and Binwise-TS achieving the lowest ECE but a slightly higher log-loss. Across both datasets, ECE does not always track improvements in proper scores: for instance, clustered Beta on IMDB has better log-loss but worse ECE than the global version. On Emotion, ECE ranks Binwise-TS above Binwise-Vector even though Binwise-Vector is strictly better in log-loss. These observations reinforce our empirical findings from tabular and image data: ECE can mis-rank calibrators relative to log-loss and Brier, supporting our choice of log-loss as the primary metric when evaluating soft, representation-based local calibration.

\begin{table}[t]
\centering

\begin{tabular}{llcccccc}
\toprule
 \textbf{Dataset}&\textbf{Method} & \textbf{K} &\textbf{$\lambda$} & \textbf{NLL $\downarrow$} & \textbf{Brier $\downarrow$} & \textbf{ECE $\downarrow$} &\textbf{Acc $\uparrow$} \\
\midrule
 IMDB&Base & -    &-& 0.503 & 0.115 & 0.141  &0.868\\
 &Histogram-Binning & -    &-& 0.355 & 0.103 & \textbf{0.012}  &0.868\\
 &Platt       & -    &-& 0.328 & 0.099 & 0.072  &0.868\\
 &Beta    & -    &-& 0.328 & 0.099 & 0.073  &0.868\\
 &TS & -    &-& 0.363 & 0.102 & 0.087  &0.868\\
 &Platt    & 4     &5& 0.327 & 0.099 & 0.073  &0.868\\
 &Beta & 4     &5& 0.324 & 0.097 & 0.075  &0.868\\
 &TS & 4  &5& 0.327 & 0.098 & 0.072  &0.868\\
 &Binwise-Platt        & -    &-& \textbf{0.300} & \textbf{0.093} & $\underline{0.013}$  &0.868\\
 &Binwise-TS  & -    &-& $\underline{0.301}$ & \textbf{0.093} & $\underline{0.013}$  &0.868\\
 &Binwise-Beta     & -    &-& \textbf{0.300} & \textbf{0.093} & \textbf{0.012}  &0.868\\
\midrule
 Emotion&Base                         & -    &-& 0.167 & 0.093 & 0.032  &\textbf{0.935}\\
 &Vector       & -    &-& 0.158 & 0.088 & 0.027  &$\underline{0.935}$\\
 &Dirichlet           & -    &-& 0.159 & 0.088 & 0.028  &0.932\\
 &TS & -    &-& 0.158 & 0.088 & 0.028  &\textbf{0.935}\\
 &Vector    & 25&1& 0.157 & $\underline{0.087}$ & 0.027  &$\underline{0.935}$\\
 &Dirichlet        & 25&1& 0.162 & $\underline{0.087}$ & 0.031  &0.932\\
 &TS & 25&1& 0.160 & $\underline{0.087}$ & 0.030  &\textbf{0.935}\\
 &Binwise-Vector        & -    &-& \textbf{0.152} & \textbf{0.086} & $\underline{0.017}$  &0.932\\
 &Binwise-TS  & -    &-& $\underline{0.156}$ & \textbf{0.086} & \textbf{0.016}  &0.934\\
\bottomrule
\end{tabular}
\caption{Calibration results on IMDB (binary sentiment) and Emotion (multi-class). We report mean NLL, Brier score, ECE, and accuracy over five random splits.
 $K$ denotes the number of clusters for clustered methods. $K=\text{'-'}$ indicates a non-clustered baseline method. Lower is better for NLL, Brier and ECE. Each entry is averaged over five random splits. 
For each method we report its best configuration (activation and shrinkage~$\lambda$) chosen by validation log-loss.
Lower is better for all metrics. Rows are unsorted, divided by dataset. Bold indicates best score and underline indicates second best.}
\label{tab:nlp}
\end{table}

\subsubsection{Robustness, Sensitivity and Ablation}
\paragraph{Tabular Data}
\subparagraph{Consistency Analysis}
A critical validation of the clustering approach is its robustness across varying model complexity. We evaluate 6 configurations: n\_estimators $\in$ \{100, 300, 1000\} and max\_depth $\in$ \{6, 8\}, representing small, medium, and large ensemble models with shallow and deeper trees\footnote{The full analysis is available in the git repository.}.
Table \ref{tab:configurations} demonstrates that clustered methods win on log-loss in all 6 configurations (100\% win rate). More remarkably, exactly 6 clustered methods appear in the top 6 log-loss ranking in all configurations except one in which 6 of 8 top methods are cluster-based, showing perfect consistency. For Brier, clustered methods occupy 6 of top 8 methods across all capacities. For AUC, the per-configuration analysis reveals 100\% clustering dominance where all the clustered methods are ranked on the top. For ECE, clustered methods occupy 6 of top 10 methods across all capacities. 
This consistency across configurations demonstrates that clustering improves model calibration in a fundamentally robust way, not dependent on specific model sizes or depths.

\subparagraph{Configuration-Specific Insights}
Across capacities (defined by n\_estimators and max\_depth), clustered calibration consistently outperforms its non-clustered baselines on both log-loss (NLL) and AUC. Aggregating over all method-capacity configurations, the best clustered variant improves NLL in 100\% of configurations with a median absolute reduction of about $1.6\times10^{-3}$ ($\approx0.55\%$ relative), and the improvement is essentially capacity-invariant (near-zero slope vs n\_estimators $\times$ depth, $R^2 \approx0.05$). In contrast, AUC gains not only occur in 100\% of configurations but also grow with capacity (positive slope with ($R^2 \approx 0.66$), yielding a median absolute increase of roughly $1.75\times10^{-3}$ ($\approx$0.21\% relative). Taken together, these results indicate that localized (cluster-aware) calibration captures subpopulation-specific miscalibration that does not wash out as the model grows. NLL gains remain stable across capacities, while added capacity tends to translate into consistent AUC improvements.

\subparagraph{Sensitivity to \#clusters}
We assessed sensitivity to the number of clusters by comparing K=4 vs K=7 on Adult and WiDS datasets across matched capacities/methods (five seeds per setting). 
Overall, AUC moves little (mean $\Delta AUC \approx +0.0035\%$, 43\% of configurations increase), while log-loss shows a small improvement (mean $\Delta NLL \approx -0.098\%$, 51\% of configurations improve). In percentage terms, raising $K$ yields modest, configuration-dependent NLL gains without a systematic AUC trade-off. Therefore, for tuning, one may pick K by validation, as we demonstrated in the image and text experiments.

\subparagraph{Ablation}
We ablated the shrinkage to global component for both SHAP and coverage clustered calibration on the Adult and WiDS datasets.  We found that AUC is essentially unchanged (mean $\Delta AUC \approx +0.0026\%$, 52\% of configurations increase) while log-loss generally worsens (mean $\Delta NLL \approx +1.29\%$, only 25\% of configurations improve). In percentage terms, removing shrinkage rarely helps and on average increases NLL while leaving AUC flat. Consistent with shrinkage acting as a stabilizer against overfitting in the calibrator. We therefore fix shrinkage $\lambda$=0.05 as the default configuration. For tuning, one may pick shrinkage $\lambda$ by validation.

\paragraph{Image and Text Datasets}
On the image and text datasets we experienced sensitivity to the number of clusters $K$  and shrinkage parameter $\lambda$. We attribute this to the sparsity and high dimensionality of the activation matrices, which appear to require larger $\lambda$ values and more fine-grained clusters. We optimized these hyper parameters using a grid search with $\lambda \in \{0.05, 1, 5, 10\}$ and $K \in \{4, 10, 25, 50\}$, with a 5-fold cross validation on the calibration data.

\subsubsection{Representation Analysis: Learned vs Data-Based Clustering}
\paragraph{Representation Comparison}
To validate methodological choices, we compare three clustering representations: two learned representations (SHAP and coverage) and one ablation using original data features (data). Table \ref{tab:performance} shows that learned representations consistently improve data-based clustering. This consistent pattern of learned representations beating data-based clustering validates that learning representations specifically captures sub-population structure relevant to model calibration. 

\subsubsection{Metric Evaluation: Why Log-Loss Matters More Than ECE}
\paragraph{Ranking Correlation Analysis}
To understand how different calibration metrics interact with predictive performance, we computed Spearman correlations between metric pairs within each model capacity (see Table\ref{tab:capacity-spearman}). log-loss shows a moderate, consistent alignment with AUC (mean $\rho \approx -0.50$ across capacities), whereas ECE is only weakly related to AUC (mean $\rho \approx -0.22$). In contrast, the LCE is positively correlated with AUC $\rho \approx 0.52$, implying that methods with better discrimination tend to have larger LCE on these experiments. With respect to accuracy, both log-loss and ECE have strong negative correlations (mean $\rho \approx -0.69$ and $-0.83$, respectively), while LCE is only moderately aligned (mean $\rho \approx -0.32$), indicating that ECE is almost a monotone transform of 0-1 loss, whereas log-loss and LCE partially decouple from pure accuracy. Finally, log-loss and Brier (both proper scores) exhibit high agreement (mean $\rho \approx 0.88$), ECE is almost perfectly coupled to Brier (mean $\rho \approx 0.94$), and LCE has only a weak association with Brier (mean $\rho \approx 0.21$). Taken together, these results support log-loss (and Brier) as principled primary calibration metrics: they are strongly aligned with proper scoring behaviour and moderately aligned with discrimination, while ECE largely duplicates Brier/accuracy and LCE can even be anti-aligned with AUC despite capturing local miscalibration.

\begin{table}[t]
\centering

\begin{tabular}{l l r}
\toprule
Metric 1 & Metric 2 & Mean Spearman $\rho$ \\
\midrule
ECE & AUC      & $-0.224$ \\
ECE & Accuracy & $-0.832$ \\
ECE & Brier    & $ 0.939$ \\
\midrule
log-loss & AUC      & $-0.502$ \\
log-loss & Accuracy & $-0.695$ \\
log-loss & Brier    & $ 0.877$ \\
\midrule
LCE & AUC      & $ 0.525$ \\
LCE & Accuracy & $-0.324$ \\
LCE & Brier    & $ 0.211$ \\
\bottomrule
\end{tabular}
\caption{Average Spearman correlations between calibration metrics (Metric~1)
and performance metrics (Metric~2), averaged over all model capacities
($n \in \{100,300,1000\}$, $d \in \{6,8\}$).}
\label{tab:capacity-spearman}
\end{table}

\subsubsection{Computational Cost and Overhead of Clustered Calibration}
We also quantified the computational overhead of clustered calibration relative to standard global calibrators. On tabular data, fitting global Beta, Platt and TS models takes between 0.0033 sec and 0.0064 sec, whereas their clustered variants over data, coverage and SHAP representations require between 0.0068 sec and 0.0149 sec, corresponding to roughly a 2-3 fold increase in calibration time (90-220\% overhead). The only notable outlier is Platt+coverage (0.1167 sec, $\approx$ 30 times slower than global Platt), but even this remains well below a tenth of a second in absolute terms. On image and text classifiers, clustered calibration over neural-network activations similarly increases optimization time by a small constant factor: Beta rises from 0.0086 sec (global) to 0.0222 sec, Platt / Vector Scaling from 1.74 sec to 4.64 sec, Dirichlet from 10.48 sec to 24.51 sec, and TS from 0.44 sec to 0.91 sec (all 2-3 fold overhead). These costs are dominated by representation extraction and clustering, which are incurred once per model: computing SHAP and coverage vectors on tabular data takes on average 0.52 sec and 1.11 sec, respectively, while k-means clustering over tabular representations adds only 0.011-0.019 sec, and clustering neural-network activations adds 0.52 sec. Overall, clustered calibration introduces only a modest constant-factor overhead over global calibration, on the order of a few additional milliseconds for tabular models and a few extra seconds for the heaviest neural-network Dirichlet setting, while providing the accuracy and discrimination gains reported in the previous sections.

\section{Conclusions and Future Work}
\subsection{Summary and Key Contributions}

This paper introduced Clustered Calibration, a principled framework for heterogeneous probability calibration via data-driven subpopulation discovery. By clustering learned representations, coverage vectors and SHAP features for tree ensembles, and hidden-layer activations for deep image and text models, the method uncovers regions of the input space that share similar predictive behavior and assigns them tailored parametric calibrators. Soft cluster memberships combined with hierarchical shrinkage toward a global mapping yield stable, context-specific calibration while preserving global calibration and avoiding overfitting in small or imbalanced clusters.

Our empirical study spans six tabular datasets, three image benchmarks and two text classification tasks, and compares Clustered Calibration against strong global and bin-wise calibration baselines. On tabular data, clustered variants of Platt/Vector, Temperature Scaling and Beta calibration consistently achieve the best or near-best negative log-likelihood (NLL) and Brier scores across model capacities, with a $100\%$ win rate in log-loss and AUC at the configuration level. These gains are systematic and they grow with model complexity, indicating that representation-based clustering becomes more beneficial as base models become more expressive. These findings establish tabular classification as the central empirical case study for Clustered Calibration: it is in this regime that representation based clustering and hierarchical shrinkage most reliably translate into improved proper scores over already strong global baselines. On image and text benchmarks, the same framework remains competitive with strong global and bin-wise methods: clustered Vector and Temperature Scaling frequently match or improve upon their global counterparts, and on several tasks achieve the best or near-best proper scores among parametric calibrators, even when bin-wise methods remain very strong baselines.

Across all domains, we observe that clustered methods typically attain ECE values that are comparable to those of the best non-clustered calibrators, indicating that exploiting subpopulation structure does not come at the expense of standard reliability metrics. At the same time, in settings where calibration is driven by soft, representation-based neighborhoods, ECE and proper scoring rules do not always induce the same ranking over methods. Our experiments, both on tabular data and on deep models, show that NLL and Brier offer a more stable and decision-theoretically grounded view of calibration quality, while ECE serves as a useful complementary diagnostic, particularly for visual inspection via reliability diagrams. Finally, we quantify computational overhead and show that Clustered Calibration adds only a small, practical cost on top of training a single global calibrator, while also providing interpretable clusters that align with meaningful subpopulations (e.g., clinically coherent risk strata in the stroke dataset).

\subsection{Limitations}

\subsubsection{Performance Gaps in Text Classification}

While Clustered Calibration demonstrated robust performance across tabular benchmarks and remained highly competitive or superior to baselines in image classification tasks, our experiments on text classification (IMDB, Emotion) revealed a different trend. In these natural language domains, bin-wise methods (such as Binwise-Vector and Binwise-TS) consistently outperformed clustered parametric calibrators. For example, on the IMDB dataset, bin-wise variants achieved a log-loss of $\approx 0.300$, whereas the best clustered variants trailed at $\approx 0.324$.

We hypothesize that this performance gap stems from specific characteristics of Transformer-based text representations with respect to calibration:

\begin{enumerate}
    \item Hyperparameter Sensitivity in Sparse Spaces: While we observed sensitivity to shrinkage ($\lambda$) and cluster count ($K$) in both image and text domains due to high dimensionality, the degradation in text suggests that the clusters found in language models are less stable or less aligned with calibration error than those in tabular and visual models, requiring further research into "calibration-aware" clustering objectives for NLP.
    \item Dominance of Confidence Signal: In the specific text tasks evaluated (Sentiment and Emotion), miscalibration appears to be primarily driven by "ambiguity" or "difficulty", which correlates strongly with the model's raw output confidence. Bin-wise methods, which partition directly on this confidence signal, capture this primary axis of variation extremely efficiently. In contrast, clustering the latent space may introduce unnecessary complexity without uncovering orthogonal "pockets" of miscalibration that are as distinct as those found in tabular or image data.
    \item Geometry of Text Embeddings: Unlike the spatially-grounded activations of CNNs or the feature-aligned splits of tree ensembles, Transformer embeddings (such as the $\verb|[CLS]|$ token from DistilBERT) often exhibit an anisotropic "cone" geometry \cite{ethayarajh2019contextual}. This structure may make standard Euclidean clustering (K-Means) less effective at isolating sub-populations with distinct calibration profiles, as semantic similarity (topic) may dominate the clustering signal over reliability signals.

\end{enumerate}

\begin{figure}[t]
  \centering
  \includegraphics[width=\linewidth]{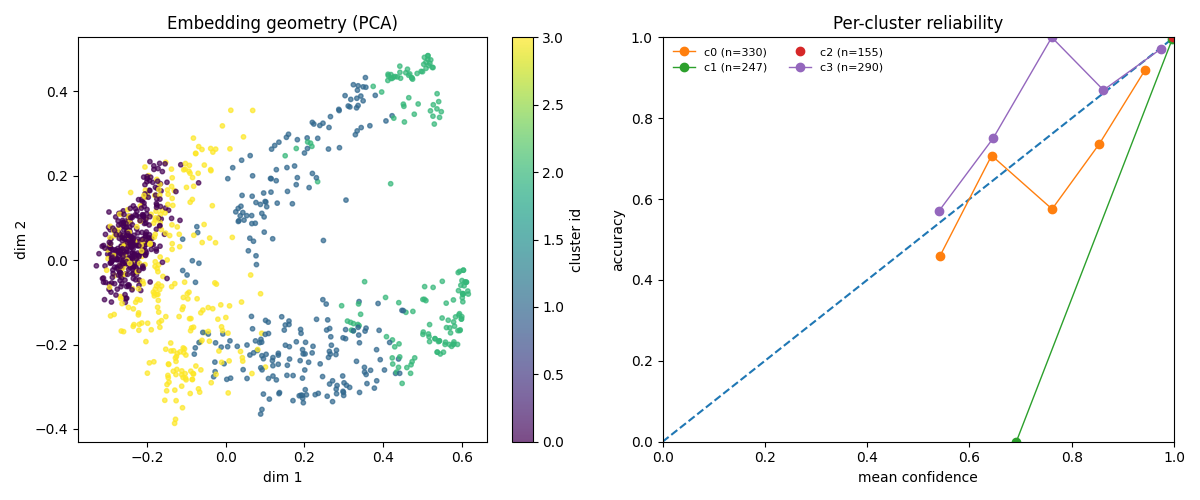}
  \caption{Stroke (tabular) cluster structure vs. calibration profiles.
  Left: 2D PCA projection of the learned representation (colored by K-Means cluster ID).
  Right: per-cluster reliability diagrams (mean accuracy vs. mean confidence, dashed line is perfect calibration).}
  \label{fig:stroke_geometry_reliability}
\end{figure}

\begin{figure}[t]
  \centering
  \includegraphics[width=\linewidth]{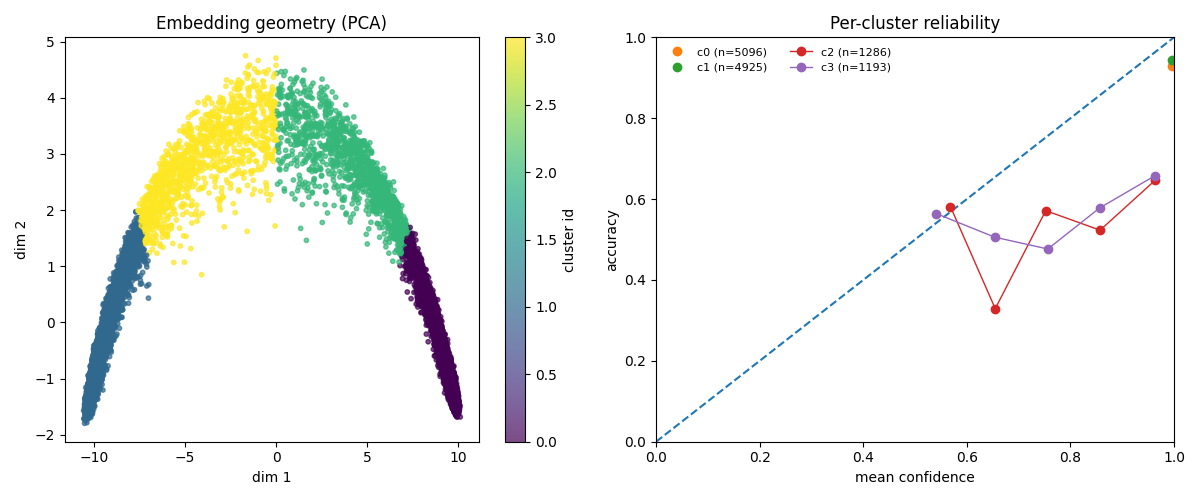}
  \caption{IMDB (text) embedding geometry and weak calibration separation.
  Left: PCA of DistilBERT [CLS] embeddings reveals a cone-like / highly anisotropic geometry with clusters arranged mainly along dominant directions.
  Right: the corresponding per-cluster reliability curves largely overlap, indicating that the induced partition captures geometric/semantic structure but does not isolate sub-populations with distinct calibration profiles.}
  \label{fig:imdb_geometry_reliability}
\end{figure}

Figures \ref{fig:stroke_geometry_reliability} and \ref{fig:imdb_geometry_reliability} contrast clustering behavior in tabular vs. text representations by pairing (i) a 2D PCA visualization of the learned embedding colored by K-Means cluster ID with (ii) per-cluster reliability diagrams. On Stroke, the coverage-based representation yields clusters whose reliability curves are more differentiated, suggesting meaningful calibration heterogeneity that cluster-specific calibrators can exploit. In contrast, IMDB’s DistilBERT [CLS] embeddings exhibit a pronounced cone-like geometry: although points cluster cleanly in PCA space, the per-cluster reliability curves largely overlap, implying that clustering is dominated by semantic/geometry structure rather than reliability signals, helping explain why clustered calibration provides limited gains on text even when geometric cluster quality (e.g., silhouette) is high.

\subsection{Future Directions}

Several promising directions emerge from this work:

\subsubsection{Integration with uncertainty quantification.}
Combining representation-based clustering with Bayesian or ensemble-based uncertainty quantification could yield joint estimates of calibrated probabilities and uncertainty intervals. Clusters may highlight regions where uncertainty is systematically under or over estimated, informing risk-aware decisions in high-stakes applications.

\subsubsection{Richer and adaptive representation schemes.}
While we demonstrated that generic learned representations (coverage vectors, SHAP, CNN activations, Transformer embeddings) already suffice to expose useful subpopulation structure, there is considerable room to explore more task-adaptive representations. Potential directions include influence-function-based embeddings, attribution maps, attention summaries or representations jointly optimized with the calibrator to better align clustering with calibration objectives.

\subsubsection{Calibration metrics and diagnostics.}
Our empirical findings suggest that proper scores (NLL, Brier) and ECE can provide partially different perspectives when calibration is performed via soft local mixtures. This motivates further work on metrics and diagnostics that (i) retain the interpretability of bin-based summaries, (ii) are robust under soft or overlapping subpopulations, and (iii) remain compatible with proper scoring rules behavior. Developing such tools could help standardize evaluation across global, bin-wise and representation-based local calibration methods and clarify when discrepancies between metrics are expected.

\subsubsection{Joint training and deployment considerations.}
Our current pipeline is post-hoc and modular: representations are extracted, clustered and then calibrated. Integrating the clustering and calibration steps more tightly with model training, for example by learning representations that are explicitly optimized for downstream calibration, or by adapting the number of clusters based on validation performance and resource constraints could further improve performance.

\subsection{Final Remarks}
Clustered Calibration demonstrates that heterogeneous calibration via learned subpopulations is both feasible and beneficial in practice. On tabular benchmarks, it delivers consistent improvements in both calibration and discrimination over strong baselines, with modest computational overhead. On image and text tasks, it remains competitive with state-of-the-art calibrators while operating on the same internal representations that modern neural networks already expose. Together with our analysis of calibration metrics and our qualitative case study on clinically meaningful clusters, these results support a general picture: exploiting the latent geometry of learned representations provides a flexible and interpretable route to improving calibrated prediction. We hope this work encourages further research on representation-aware calibration, stronger evaluation practices, and cluster-based tools that help practitioners reason about model reliability at the level of the subpopulations that matter in downstream decisions.

\section{Declaration}
\textbf{Funding.} his research received no specific grant from any funding agency in the public, commercial, or not-for-profit sectors.

\textbf{Conflict of interest.} The authors declare that they have no conflict of interest.

\textbf{Ethical approval.} This article does not contain any studies with human participants or animals performed by any of the authors beyond the use of fully de-identified benchmark datasets.

\textbf{Use of AI tools.} Language-editing and programming code assistance was provided by large language models. All content was reviewed and verified by the authors.

% Acknowledgements and Disclosure of Funding should go at the end, before appendices and references

% Manual newpage inserted to improve layout of sample file - not
% needed in general before appendices/bibliography.

\newpage

\begin{appendices}

\section{ECE mis-ranking under soft mixtures: full proof}
\label{app:ece_misrank_full}

\subsection{Proposition (existence)}

For any fixed finite partition \(\mathcal{B}\) of \([0,1]\) into probability bins, there exist a binary classification problem and two predictors: a constant (global) predictor \(P_{\mathrm{glob}}\) and a local/clustered predictor \(P_{\mathrm{soft}}\) s.t.
\[
\mathrm{NLL}(P_{\mathrm{soft}}) < \mathrm{NLL}(P_{\mathrm{glob}})
\]
but
\[
\mathrm{ECE}_{\mathcal{B}}(P_{\mathrm{soft}}) > \mathrm{ECE}_{\mathcal{B}}(P_{\mathrm{glob}}).
\]

\subsubsection{Definitions used}
\begin{itemize}
    \item For \(p\in(0,1)\) and a predicted probability \(q\in(0,1)\), the per-example negative log-likelihood (cross-entropy) is
  \[
  \mathrm{CE}(p,q) = -p\log q -(1-p)\log(1-q).
  \]
  \item ECE under a finite partition \(\mathcal{B}={B_1,\ldots,B_K}\) of \([0,1]\) is
  \[
  \mathrm{ECE}_{\mathcal{B}}(P)=\sum_{k=1}^K w_k\big|\overline{\mathrm{acc}}(B_k) - \overline{\mathrm{conf}}(B_k)\big|,
  \]
where \(w_k\) is the fraction of test points whose predicted probability falls in bin \(B_k\),  \(\overline{\mathrm{conf}}(B_k)\) is the mean predicted probability in that bin, and \(\overline{\mathrm{acc}}(B_k)\) is the empirical positive rate in that bin.
\end{itemize}

\subsection{Construction}
\subsubsection{Step 1 (choose two bins and anchor points).}
Because \(\mathcal{B}\) is finite, it has at least two bins. Pick one “low” bin \(B^{-}\) and one “high” bin \(B^{+}\). Choose any numbers
\[
\alpha \in B^{-}, \beta \in B^{+} \quad \text{with } 0<\alpha<\beta<1.
\]
Let \(c := (\alpha+\beta)/2\) be their midpoint. 

\subsubsection{Step 2 (data distribution).}
Let \(X\in\{A,B\}\) with \(\Pr(X=A)=\Pr(X=B)=\tfrac12\).
Conditional on $X$:
\[
Y\mid X=A \sim \mathrm{Bern}(p_A), \qquad Y\mid X=B \sim \mathrm{Bern}(p_B),
\]
where we set
\[
p_A := \beta+\delta,\qquad p_B := \alpha-\delta
\]
for some small \(\delta>0\) chosen so that \(p_A,p_B\in(0,1)\).
The global prevalence is \(\bar p := \Pr(Y{=}1)=\tfrac12(p_A+p_B)=\tfrac12(\beta+\delta+\alpha-\delta)=c\).

\subsubsection{Step 3 (predictors).}
\begin{itemize}
    \item Global predictor: \(P_{\mathrm{glob}}(x)\equiv c\) for all $x$.
    \item Local/clustered predictor: predict \(\beta\) on region \(A\) and \(\alpha\) on region \(B\).
\end{itemize}

We now compare NLL and ECE for these two predictors.

\subsection{Part A} 
\[ \bm {\mathrm{ECE}_{\mathcal{B}}(P_{\mathrm{soft}}) > \mathrm{ECE}_{\mathcal{B}}(P_{\mathrm{glob}})}\]

\begin{itemize}
    \item Under \(P_{\mathrm{glob}}\), all predictions equal \(c\), thus all points fall into the single bin \(B\) containing \(c\). The empirical accuracy in that bin equals the global prevalence \( \bar p = c\). Hence the bin’s contribution to ECE is \(|\bar p - c|=0\), and therefore
  \[
  \mathrm{ECE}_{\mathcal{B}}(P_{\mathrm{glob}})=0.
  \]

\item Under \(P_{\mathrm{soft}}\), all \(A\)-points predict \(\beta\in B^{+}\) and all \(B\)-points predict \(\alpha\in B^{-}\).
The mean confidence and empirical accuracy in those bins are:
  \[
  \text{in } B^{+}:\ (\overline{\mathrm{conf}},\overline{\mathrm{acc}})=(\beta,p_A=\beta+\delta),\quad
\]
\[  \text{in } B^{-}:\ (\overline{\mathrm{conf}},\overline{\mathrm{acc}})=(\alpha,p_B=\alpha-\delta).
 \]
  Each bin has weight \(w=\tfrac12\). Thus
  \[
  \mathrm{ECE}_{\mathcal{B}}(P_{\mathrm{soft}})
  = \tfrac12\big|(\beta+\delta)-\beta\big|+\tfrac12\big|(\alpha-\delta)-\alpha\big|
  = \tfrac12\delta + \tfrac12\delta = \delta >0.
\]

\end{itemize}

Therefore \(\mathrm{ECE}_{\mathcal{B}}(P_{\mathrm{soft}}) > \mathrm{ECE}_{\mathcal{B}}(P_{\mathrm{glob}})\).

\subsection{Part B}  \[\bm {\mathrm{NLL}(P_{\mathrm{soft}}) < \mathrm{NLL}(P_{\mathrm{glob}})}\]
Write \(\mathrm{CE}(p,q) = -p\log q -(1-p)\log(1-q)\). For our mixture with equal mass in \(A\) and \(B\),
\begin{itemize}
    \item Global predictor:
  \[
  \mathrm{NLL}(P_{\mathrm{glob}}) = \tfrac12\mathrm{CE}(p_A,c)+\tfrac12\mathrm{CE}(p_B,c).
  \]
\item  Local predictor:
  \[
  \mathrm{NLL}(P_{\mathrm{soft}}) = \tfrac12\mathrm{CE}(p_A,\beta)+\tfrac12\mathrm{CE}(p_B,\alpha).
  \]
\end{itemize}

Now observe the distances:
\begin{itemize}
    \item For region \(A\): \(p_A=\beta+\delta\). Distances to the two competing predictions are:  
\[
  |p_A-\beta|=\delta,\qquad |p_A-c|=|(\beta+\delta)-\tfrac12(\alpha+\beta)|=\tfrac12(\beta-\alpha)+\delta.
  \]
  Since \(\beta>\alpha\) and \(\delta>0\), we have \(|p_A-\beta|<|p_A-c|\). By the monotonicity just noted,
  \[
  \mathrm{CE}(p_A,\beta) < \mathrm{CE}(p_A,c).
  \]

\item  For region \(B\): \(p_B=\alpha-\delta\). Distances are
  \[
  |p_B-\alpha|=\delta,\qquad |p_B-c|=|\ (\alpha-\delta)-\tfrac12(\alpha+\beta)\ |=\tfrac12(\beta-\alpha)+\delta.
  \]
  Again \(|p_B-\alpha|<|p_B-c|\), hence
  \[
  \mathrm{CE}(p_B,\alpha) < \mathrm{CE}(p_B,c).
  \]
\end{itemize}

Averaging the two strict inequalities yields
\[
\mathrm{NLL}(P_{\mathrm{soft}})
= \tfrac12\mathrm{CE}(p_A,\beta)+\tfrac12\mathrm{CE}(p_B,\alpha)
< \tfrac12\mathrm{CE}(p_A,c)+\tfrac12\mathrm{CE}(p_B,c)
= \mathrm{NLL}(P_{\mathrm{glob}}).
\]

This completes the proof.

\section{Theoretical Properties of Cluster-Based Calibration}
\label{app:theory-cluster-calib}

In this section we formalize a simple population-level guarantee for cluster-based calibration under the Brier score. We show that if the calibrator is allowed to use (i) only the base score $S$ or (ii) the base score $S$ together with a cluster-based representation $R$, then the population-optimal cluster-aware calibrator is always at least as close to the Bayes predictor as the population-optimal global calibrator. This guarantee is preserved under any convex $\lambda$-shrinkage toward the global solution.

\subsection{Setup}
We consider binary classification with input $X \in \mathcal{X}$ and label
$Y \in \{0,1\}$. Let
\[
\eta(X) := \mathbb{P}(Y=1 \mid X)
\]
denote the Bayes conditional probability. A base classifier produces a score $S = f(X) \in \mathbb{R}$. In addition, we assume access to a learned representation $Z = \phi(X)$ and to cluster information $R = C(Z)$ derived from it. The variable $R$ may encode either hard cluster assignments, $R \in \{1,\dots,K\}$, or soft cluster-membership weights $R = (R_1,\dots,R_K)$ with $R_k \in [0,1]$ and $\sum_k R_k = 1$. In all cases $R$ is a (measurable) function of $X$.

We write $\mathcal{F}_0 := \sigma(S)$ for the $\sigma$-algebra generated by $S$
and $\mathcal{F}_1 := \sigma(S,R)$ for the $\sigma$-algebra generated by $(S,R)$. Since the pair $(S,R)$ contains at least as much information as the score $S$ alone, the $\sigma$-algebra generated by $S$ is contained in the $\sigma$-algebra generated by $(S,R)$, i.e. $\sigma(S) \subseteq \sigma(S,R)$. I.e. $\mathcal{F}_0 \subseteq \mathcal{F}_1$. We evaluate probabilistic predictions $g(X) \in [0,1]$ using the Brier score:
\[
\mathcal{R}(g) := \mathbb{E}\bigl[(Y - g(X))^2\bigr].
\]
Using the identity
\[
\mathcal{R}(g)
= \mathbb{E}\bigl[(\eta(X) - g(X))^2\bigr]
  + \mathbb{E}\bigl[\eta(X)(1-\eta(X))\bigr],
\]
minimizing $\mathcal{R}(g)$ is equivalent to minimizing
\[
\tilde{\mathcal{R}}(g) := \mathbb{E}\bigl[(\eta(X) - g(X))^2\bigr],
\]
the squared $L^2$-distance to the Bayes predictor.

We define the population-optimal global and cluster-aware calibrators as
\begin{align*}
g_0^\star(S) &:= \mathbb{E}[\,Y \mid \mathcal{F}_0\,] = \mathbb{E}[\,Y \mid S\,], \\
g_1^\star(S,R) &:= \mathbb{E}[\,Y \mid \mathcal{F}_1\,] = \mathbb{E}[\,Y \mid S,R\,].
\end{align*}
Both are conditional expectations, i.e., $L^2$-projections of $Y$ onto the spaces of $\mathcal{F}_0$- and $\mathcal{F}_1$-measurable functions, respectively.

\subsection{More information under Brier risk}

\begin{theorem}[Cluster-aware calibrator is at least as close to Bayes as a global calibrator]
\label{thm:cluster-dominates-global}
Under the setup above,
\begin{equation}
\label{eq:cluster-dominates-global}
\tilde{\mathcal{R}}(g_1^\star)
= \mathbb{E}\bigl[(\eta(X) - g_1^\star(S,R))^2\bigr]
\le
\mathbb{E}\bigl[(\eta(X) - g_0^\star(S))^2\bigr]
= \tilde{\mathcal{R}}(g_0^\star).
\end{equation}
Equivalently,
\[
\mathcal{R}(g_1^\star) \le \mathcal{R}(g_0^\star),
\]
so the population-optimal cluster-aware calibrator is never worse than the population-optimal global calibrator under the Brier score.
\end{theorem}

\begin{proof}
$g_0^\star = \mathbb{E}[Y \mid \mathcal{F}_0]$ and $g_1^\star = \mathbb{E}[Y \mid \mathcal{F}_1]$ are conditional expectations. They are the unique minimizers of
\[
g \mapsto \mathbb{E}\bigl[(Y - g(X))^2\bigr]
\]
over all $\mathcal{F}_0$ and $\mathcal{F}_1$ respectively.
Since $\mathcal{F}_0 \subseteq \mathcal{F}_1$, every $\mathcal{F}_0$-measurable function is also $\mathcal{F}_1$-measurable. Therefore
\[
\mathcal{R}(g_1^\star)
= \inf_{g: \ \mathcal{F}_1\text{-measurable}} \mathbb{E}\bigl[(Y-g(X))^2\bigr]
\le
\inf_{g: \ \mathcal{F}_0\text{-measurable}} \mathbb{E}\bigl[(Y-g(X))^2\bigr]
= \mathcal{R}(g_0^\star).
\]
Using the decomposition
\[
\mathcal{R}(g)
= \tilde{\mathcal{R}}(g)
  + \mathbb{E}\bigl[\eta(X)(1-\eta(X))\bigr],
\]
where the second term does not depend on $g$, the inequality
$\mathcal{R}(g_1^\star) \le \mathcal{R}(g_0^\star)$ is equivalent to $\tilde{\mathcal{R}}(g_1^\star) \le \tilde{\mathcal{R}}(g_0^\star)$, which is  \eqref{eq:cluster-dominates-global}.
\end{proof}
\begin{remark}
At the population level, the best cluster-aware calibrator is $g_1^\star(S,R) = \mathbb{E}[Y|S,R]$, which cannot be worse than the best global calibrator $g_0^\star(S) = \mathbb{E}[Y|S]$ under the Brier score (Theorem \ref{thm:cluster-dominates-global}). In practice, we estimate these functions using finite calibration sets and parametric families. To reduce variance in small clusters, we add an $\ell_2$ penalty that shrinks each cluster-specific parameter vector $\theta_k$ toward the global parameters $\theta_0$, as in Eq. \ref{eq:CCL-Loss}.
\end{remark}

\newpage

%%=============================================%%
%% For submissions to Nature Portfolio Journals %%
%% please use the heading ``Extended Data''.   %%
%%=============================================%%

%%=============================================================%%
%% Sample for another appendix section			       %%
%%=============================================================%%

%% \section{Example of another appendix section}\label{secA2}%
%% Appendices may be used for helpful, supporting or essential material that would otherwise 
%% clutter, break up or be distracting to the text. Appendices can consist of sections, figures, 
%% tables and equations etc.

\end{appendices}

%%===========================================================================================%%
%% If you are submitting to one of the Nature Portfolio journals, using the eJP submission   %%
%% system, please include the references within the manuscript file itself. You may do this  %%
%% by copying the reference list from your .bbl file, paste it into the main manuscript .tex %%
%% file, and delete the associated \verb+\bibliography+ commands.                            %%
%%===========================================================================================%%

\bibliography{sn-article}% common bib file
%% if required, the content of .bbl file can be included here once bbl is generated
%%\input sn-article.bbl

\end{document}